# A Survey of Large Language Models in Medicine: Progress, Application, and Challenge

Hongjian Zhou[1,*], Fenglin Liu[1,*,†] Boyang Gu[2,*], Xinyu Zou[3,*], Jinfa Huang[4,*], Jinge Wu[5,*], Yiru Li[6], Sam Chen[7], Peilin Zhou[8], Junling Liu[9], Yining Hua[10], Chengfeng Mao[11], Chenyu You[12], Xian Wu[13], Yefeng Zheng[13], Lei Clifton[1], Zheng Li[14,†], Jiebo Luo[4,†], David Clifton[1,15,†]

[*] Core Contributors, ordered by a coin toss. † Corresponding Authors.
[1]University of Oxford, [2]Imperial College London, [3]University of Waterloo,
[4]University of Rochester, [5]University College London, [6]Western University,
[7]University of Georgia, [8]Hong Kong University of Science and Technology (Guangzhou),
[9]Alibaba, [10]Harvard T.H. Chan School of Public Health, [11]Massachusetts Institute of Technology,
[12]Yale University, [13]Tencent, [14]Amazon, [15]Oxford-Suzhou Centre for Advanced Research
{fenglin.liu,david.clifton}@eng.ox.ac.uk, amzzhe@amazon.com, jluo@cs.rochester.edu

## ABSTRACT

Large language models (LLMs), such as ChatGPT, have received substantial attention due to their capabilities for understanding and generating human language. While there has been a burgeoning trend in research focusing on the employment of LLMs in supporting different medical tasks (e.g., enhancing clinical diagnostics and providing medical education), a comprehensive review of these efforts, particularly their development, practical applications, and outcomes in medicine, remains scarce. Therefore, this review aims to provide a detailed overview of the development and deployment of LLMs in medicine, including the challenges and opportunities they face. In terms of development, we provide a detailed introduction to the principles of existing medical LLMs, including their basic model structures, number of parameters, and sources and scales of data used for model development. It serves as a guide for practitioners in developing medical LLMs tailored to their specific needs. In terms of deployment, we offer a comparison of the performance of different LLMs across various medical tasks, and further compare them with state-of-the-art lightweight models, aiming to provide a clear understanding of the distinct advantages and limitations of LLMs in medicine. Overall, in this review, we address the following study questions: 1) What are the practices for developing medical LLMs? 2) How to measure the medical task performance of LLMs in a medical setting? 3) How have medical LLMs been employed in real-world practice? 4) What challenges arise from the use of medical LLMs? and 5) How to more effectively develop and deploy medical LLMs? By answering these questions, this review aims to provide insights into the opportunities and challenges of LLMs in medicine and serve as a practical resource for constructing effective medical LLMs. We also maintain a regularly updated list of practical guides on medical LLMs at: https://github.com/AI-in-Health/MedLLMsPracticalGuide.

### BOX: Key points

- Existing medical LLMs, ranging from 110 million to 520 billion parameters, are mainly developed through pre-training, fine-tuning, and prompting methods, utilizing large-scale medical corpora from diverse sources.

- Most existing works evaluate their performance on exam-style QA tasks. A reasonable combination of different fine-tuning and prompting methods enables LLMs to achieve comparable or even better results than experts.

- LLMs' poor performance in non-QA tasks without pre-set options, which are common in clinical practice, indicates a considerable need for advancement before LLMs can be integrated into the actual clinical decision-making process.

- Adapting medical LLMs to various clinical applications has received increasing research interest. However, large-scale clinical trials specifically targeting existing medical LLMs are currently missing.

- Addressing the challenges and future directions—regarding mitigating hallucinations; establishing robust data, benchmarks, metrics; and addressing ethical, safety, and regulatory concerns, through interdisciplinary collaborations—is important to accelerate the integration of LLMs into the clinic.

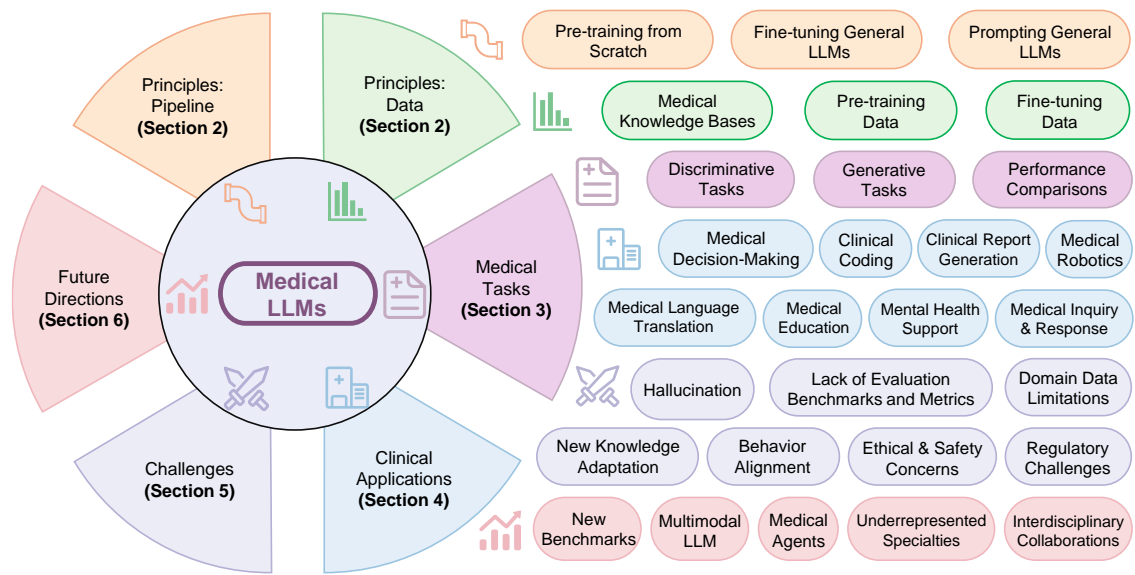

**Figure 1.** An overview of the practical guides for medical large language models.

## 1 Introduction

The recently emerged general large language models (LLMs)[1,2], such as PaLM[3], LLaMA[4,5], GPT-series[6,7], and ChatGLM[8,9], have advanced the state-of-the-art in various natural language processing (NLP) tasks, including text generation, text summarization, and question answering. Inspired by these successes, several endeavors have been made to adapt general LLMs to the medicine domain, leading to the emergence of medical LLMs[10,11]. For example, based on PaLM[3] and GPT-4[7], MedPaLM-2[11] and MedPrompt[12] have respectively achieved a competitive accuracy of 86.5 and 90.2 compared to human experts (87.0[13]) in the United States Medical Licensing Examination (USMLE)[14]. In particular, based on publicly available general LLMs (e.g. LLaMA[4,5]), a wide range of medical LLMs, including ChatDoctor[15], MedAlpaca[16], PMC-LLaMA[13], BenTsao[17], and Clinical Camel[18], have been introduced. As a result, medical LLMs have gained growing research interests in assisting medical professionals to improve patient care[19,20].

Although existing medical LLMs have achieved promising results, there are some key issues in their development and application that need to be addressed. First, many of these models primarily focus on medical dialogue and medical question-answering tasks, but their practical utility in clinical practice is often overlooked[19]. Recent research and reviews[19,21,22] have begun to explore the potential of medical LLMs in different clinical scenarios, including Electronic Health Records (EHRs)[23], discharge summary generation[20], health education[24], and care planning[11]. However, they primarily focus on presenting clinical applications of LLMs, especially online commercial LLMs like ChatGPT (including GPT-3.5 and GPT-4[7]), without providing practical guidelines for the development of medical LLMs. Besides, they mainly perform case studies to conduct the human evaluation on a small number of samples, thus lacking evaluation datasets for assessing model performance in clinical scenarios. Second, most existing medical LLMs report their performances mainly on answering medical questions, neglecting other biomedical domains, such as medical language understanding and generation. These research gaps motivate this review which offers a comprehensive review of the development of LLMs and their applications in medicine. We aim to cover topics on existing medical LLMs, various medical tasks, clinical applications, and arising challenges.

As shown in Figure 1, this review seeks to answer the following questions. **Section 2**: What are LLMs? How can medical LLMs be effectively built? **Section 3**: How are the current medical LLMs evaluated? What capabilities do medical LLMs offer beyond traditional models? **Section 4**: How should medical LLMs be applied in clinical settings? **Section 5**: What challenges should be addressed when implementing medical LLMs in clinical practice? **Section 6**: How can we optimize the construction of medical LLMs to enhance their applicability in clinical settings, ultimately contributing to medicine and creating a positive societal impact?

For the first question, we analyze the foundational principles underpinning current medical LLMs, providing detailed descriptions of their architecture, parameter scales, and the datasets used during their development. This exposition aims to serve as a valuable resource for researchers and clinicians designing medical LLMs tailored to specific requirements, such as computational constraints, data privacy concerns, and the integration of local knowledge bases. For the second question, we evaluate the performance of medical LLMs across ten biomedical NLP tasks, encompassing both discriminative and generative tasks. This comparative analysis elucidates how these models outperform traditional AI approaches, offering insights into the specific capabilities that render LLMs effective in clinical environments. The third question, the practical deployment of medical LLMs in clinical settings, is explored through the development of guidelines tailored for seven distinct clinical application scenarios. This section outlines practical implementations, emphasizing specific functionalities of medical LLMs that are leveraged in each scenario. The fourth question emphasizes addressing the challenges associated with the clinical deployment of medical LLMs, such as the risk of generating factually inaccurate yet plausible outputs (hallucination), and the ethical, legal, and safety implications. Citing recent studies, we argue for a comprehensive evaluation framework that assesses the trustworthiness of medical LLMs to ensure their responsible and effective utilization in healthcare. For the last question, we propose future research directions to advance the medical LLMs field. This includes fostering interdisciplinary collaboration between AI specialists and medical professionals, advocating for a 'doctor-in-the-loop' approach, and emphasizing human-centered design principles.

By establishing robust training data, benchmarks, metrics, and deployment strategies through co-development efforts, we aim to accelerate the responsible and efficacious integration of medical LLMs into clinical practice. This study therefore seeks to stimulate continued research and development in this interdisciplinary field, with the objective of realizing the profound potential of medical LLMs in enhancing clinical practice and advancing medical science for the betterment of society.

## BOX 1: Background of Large Language Models (LLMs)

The impressive performance of LLMs can be attributed to Transformer-based language models, large-scale pre-training, and scaling laws.

**Language Models** A language model [25,26,27] is a probabilistic model that models the joint probability distribution of tokens (meaningful units of text, such as words or subwords or morphemes) in a sequence, i.e., the probabilities of how words and phrases are used in sequences. Therefore, it can predict the likelihood of a sequence of tokens given the previous tokens, which can be used to predict the next token in a sequence or to generate new sequences.

**The Transformer architecture** The recurrent neural network (RNN) [28,26] has been widely used for language modeling by processing tokens sequentially and maintaining a vector named hidden state that encodes the context of previous tokens. Nonetheless, sequential processing makes it unsuitable for parallel training and limits its ability to capture long-range dependencies, making it computationally expensive and hindering its learning ability for long sequences. The strength of the Transformer [29] lies in its fully attentive mechanism, which relies exclusively on the attention mechanism and eliminates the need for recurrence. When processing each token, the attention mechanism computes a weighted sum of the other input tokens, where the weights are determined by the relevance between each input token and the current token. It allows the model to adaptively focus on different parts of the sequence to effectively learn the joint probability distribution of tokens. Therefore, Transformer not only enables efficient modeling of long-text but also allows highly paralleled training [30], thus reducing training costs. They make the Transformer highly scalable, and therefore it is efficient to obtain LLMs through the large-scale pre-training strategy.

**Large-scale Pre-training** The LLMs are trained on massive corpora of unlabeled texts (e.g., CommonCrawl, Wiki, and Books) to learn rich linguistic knowledge and language patterns. The common training objectives are *masked language modeling (MLM)* and *next token prediction (NTP)*. In MLM, a portion of the input text is masked, and the model is tasked with predicting the masked text based on the remaining unmasked context, encouraging the model to capture the semantic and syntactic relationships between tokens [30]; NTP is another common training objective, where the model is required to predict the next token in a sequence given the previous tokens. It helps the model to predict the next token [6].

**Scaling Laws** LLMs are the scaled-up versions of Transformer architecture [29] with increased numbers of Transformer layers, model parameters, and volume of pre-training data. The "scaling laws" [31,32] predict how much improvement can be expected in a model's performance as its size increases (in terms of parameters, layers, data, or the amount of training computed). The scaling laws proposed by OpenAI [31] show that to achieve optimal model performance, the budget allocation for model size should be larger than the data.

The scaling laws proposed by Google DeepMind [32] show that both model and data sizes should be increased in equal scales. The scaling laws guide researchers to allocate resources and anticipate the benefits of scaling models.

**General Large Language Models** Existing general LLMs can be divided into three categories based on their architecture (Table 1).

*Encoder-only LLMs* consisting of a stack of Transformer encoder layers, employ a bidirectional training strategy that allows them to integrate context from both the left and the right of a given token in the input sequence. This bi-directionality enables the models to achieve a deep understanding of the input sentences [30]. Therefore, encoder-only LLMs are particularly suitable for language understanding tasks (e.g., sentiment analysis document classification) where the full context of the input is essential for accurate predictions. BERT [30] and DeBERTa [33] are the representative encoder-only LLMs.

*Decoder-only LLMs* utilize a stack of Transformer decoder layers and are characterized by their uni-directional (left-to-right) processing of text, enabling them to generate language sequentially. This architecture is trained unidirectionally using the next token prediction training objective to predict the next token in a sequence, given all the previous tokens. After training, the decoder-only LLMs generate sequences autoregressively (i.e. token-by-token). The examples are the GPT-series developed by OpenAI [6,7], the LLaMA-series developed by Meta [4,5], and the PaLM [3] and Bard (Gemini) [34] developed by Google. Based on the LLaMA model, Alpaca [35] is fine-tuned with 52k self-instructed data supervision. In addition, Baichuan [36] is trained on approximately 1.2 trillion tokens that support bilingual communication in Chinese and English. These LLMs have been used successfully in language generation.

*Encoder-decoder LLMs* are designed to simultaneously process input sequences and generate output sequences. They consist of a stack of bidirectional Transformer encoder layers followed by a stack of unidirectional Transformer decoder layers. The encoder processes and understands the input sequences, while the decoder generates the output sequences [8,9,37]. Representative examples of encoder-decoder LLMs include Flan-T5 [38], and ChatGLM [8,9]. Specifically, ChatGLM [8,9] has 6.2B parameters and is a conversational open-source LLM specially optimized for Chinese to support Chinese-English bilingual question-answering.

**Table 1.** Summary of existing general (large) language models, their underlying structures, numbers of parameters, and datasets used for model training. Column "# params" shows the number of parameters, M: million, B: billion.

| Domains | Model Structures | Models | # Params | Pre-train Data Scale |
|---|---|---|---|---|
| | Encoder-only | BERT [30] | 110M/340M | 3.3B tokens |
| | | RoBERTa [39] | 355M | 161GB |
| | | DeBERTa [33] | 1.5B | 160GB |
| General-domain (Large) Language Models | Decoder-only | GPT-2 [40] | 1.5B | 40GB |
| | | Vicuna [41] | 7B/13B | LLaMA + 70K dialogues |
| | | Alpaca [35] | 7B/13B | LLaMA+ 52K IFT |
| | | Mistral [42] | 7B | - |
| | | LLaMA [4] | 7B/13B/33B/65B | 1.4T tokens |
| | | LLaMA-2 [5] | 7B/13B/34B/70B | 2T tokens |
| | | LLaMA-3 [43] | 8B/70B | 15T tokens |
| | | GPT-3 [6] | 6.7B/13B/175B | 300B tokens |
| | | Qwen [44] | 1.8B/7B/14B/72B | 3T tokens |
| | | PaLM [3] | 8B/62B/540B | 780B tokens |
| | | FLAN-PaLM [37] | 540B | - |
| | | Gemini (Bard) [34] | - | - |
| | | GPT-3.5 [45] | - | - |
| | | GPT-4 [7] | - | - |
| | | Claude-3 [46] | - | - |
| | Encoder-Decoder | BART [47] | 140M/400M | 160GB |
| | | ChatGLM [8,9] | 6.2B | 1T tokens |
| | | T5 [38] | 11B | 1T tokens |
| | | FLAN-T5 [37] | 3B/11B | 780B tokens |
| | | UL2 [48] | 19.5B | 1T tokens |
| | | GLM [9] | 130B | 400B tokens |

## 2 The Principles of Medical Large Language Models

Box 1 and Table 1 briefly introduce the background of general LLMs [1], e.g., GPT-4 [7]. Table 2 summarizes the currently available medical LLMs according to their model development. Existing medical LLMs are mainly pre-trained from scratch, fine-tuned from existing general LLMs, or directly obtained through prompting to align the general LLMs to the medical domain. Therefore, we introduce the principles of medical LLMs in terms of these three methods: pre-training, fine-tuning, and prompting. Meanwhile, we further summarize the medical LLMs according to their model architectures in Figure 2.

**Table 2.** Summary of existing medical-domain LLMs, in terms of their model development, the number of parameters (# params), the scale of pre-training/fine-tuning data, and the data source. M: million, B: billion.

| Domains | Model Development | Models | # Params | Data Scale | Data Source |
|---|---|---|---|---|---|
| Medical-domain LLMs (Sec. 2) | Pre-training (Sec. 2.1) | BioBERT[49] | 110M | 18B tokens | PubMed[50]+PMC[51] |
| | | PubMedBERT[52] | 110M/340M | 3.2B tokens | PubMed[50]+PMC[51] |
| | | SciBERT[53] | 110M | 3.17B tokens | Literature[54] |
| | | NYUTron[55] | 110M | 7.25M notes, 4.1B tokens | NYU Notes[55] |
| | | ClinicalBERT[56] | 110M | 112k clinical notes | MIMIC-III[57] |
| | | BioM-ELECTRA[58] | 110M/335M | - | PubMed[50] |
| | | BioMed-RoBERTa[59] | 125M | 7.55B tokens | S2ORC[60] |
| | | BioLinkBERT[61] | 110M/340M | 21GB | PubMed[50] |
| | | BlueBERT[62,63,64] | 110M/340M | >4.5B tokens | PubMed[50]+MIMIC-III[57] |
| | | SciFive[65] | 220M/770M | - | PubMed[50]+PMC[51] |
| | | ClinicalT5[66] | 220M/770M | 2M clinical notes | MIMIC-III[57] |
| | | MedCPT[67] | 330M | 255M articles | PubMed[50] |
| | | DRAGON[68] | 360M | 6GB | BookCorpus[69] |
| | | BioGPT[70] | 1.5B | 15M articles | PubMed[50] |
| | | BioMedLM[71] | 2.7B | 110GB | Pile[72] |
| | | OphGLM[73] | 6.2B | 20k dialogues | MedDialog[74] |
| | | GatorTron[23] | 8.9B | >82B tokens+6B tokens 2.5B tokens+0.5B tokens | EHRs[23]+PubMed[50] Wiki+MIMIC-III[57] |
| | | GatorTronGPT[75] | 5B/20B | 277B tokens | EHRs[75] |
| | Fine-tuning (Sec. 2.2) | DoctorGLM[76] | 6.2B | 323MB dialogues | CMD.[77] |
| | | BianQue[78] | 6.2B | 2.4M dialogues | BianQueCorpus[78] |
| | | ClinicalGPT[79] | 7B | 96k EHRs + 100k dialogues 192 medical QA | MD-EHR[79]+MedDialog[74] VariousMedQA[14] |
| | | Qilin-Med[80] | 7B | 3GB | ChiMed[80] |
| | | ChatDoctor[15] | 7B | 110k dialogues | HealthCareMagic[81]+iCliniq[82] |
| | | BenTsao[17] | 7B | 8k instructions | CMeKG-8K[83] |
| | | HuatuoGPT[84] | 7B | 226k instructions&dialogues | Hybrid SFT[84] |
| | | Baize-healthcare[85] | 7B | 101K dialogues | Quora+MedQuAD[86] |
| | | BioMedGPT[87] | 10B | >26B tokens | S2ORC[60] |
| | | MedAlpaca[16] | 7B/13B | 160k medical QA | Medical Meadow[16] |
| | | AlpaCare[88] | 7B/13B | 52k instructions | MedInstruct-52k[88] |
| | | Zhongjing[89] | 13B | 70k dialogues | CMtMedQA[89] |
| | | PMC-LLaMA[13] | 13B | 79.2B tokens | Books+Literature[60]+MedC-I[13] |
| | | CPLLM[90] | 13B | 109k EHRs | eICU-CRD[91]+MIMIC-IV[92] |
| | | Med42[93] | 7B/70B | 250M tokens | PubMed[50] + MedQA[14] + OpenOrca |
| | | MEDITRON[94,95] | 7B/70B | 48.1B tokens | PubMed[50]+Guidelines[94] |
| | | OpenBioLLM[96] | 8B/70B | - | - |
| | | MedLlama3-v20[97] | 8B/70B | - | - |
| | | Clinical Camel[18] | 13B/70B | 70k dialogues+100k articles 4k medical QA | ShareGPT[98]+PubMed[50] MedQA[14] |
| | | MedPaLM-2[11] | 340B | 193k medical QA | MultiMedQA[11] |
| | | Med-Flamingo[99] | - | 600k pairs | Multimodal Textbook[99]+PMC-OA[99] VQA-RAD[100]+PathVQA[101] |
| | | LLaVA-Med[102] | - | 660k pairs | PMC-15M[102]+VQA-RAD[100] SLAKE[103]+PathVQA[101] |
| | | MAIRA-1[104] | - | 337k pairs | MIMIC-CXR[105] |
| | | RadFM[106] | - | 32M pairs | MedMD[106] |
| | | Med-Gemini[107,108] | - | - | MedQA-R&RS[108]++MultiMedQA[11] +MIMIC-III[57]+MultiMedBench[109] |
| | Prompting (Sec. 2.3) | CodeX[110] | GPT-3.5 / LLaMA-2 | Chain-of-Thought (CoT)[111] | - |
| | | DeID-GPT[112] | ChatGPT / GPT-4 | Chain-of-Thought (CoT)[111] | - |
| | | ChatCAD[113] | ChatGPT | In-Context Learning (ICL) | - |
| | | Dr. Knows[114] | ChatGPT | ICL | UMLS[115] |
| | | MedPaLM[10] | PaLM (540B) | CoT & ICL | MultiMedQA[11] |
| | | MedPrompt[12] | GPT-4 | CoT & ICL[111] | - |
| | | Chat-Orthopedist[116] | ChatGPT | Retrieval-Augmented Generation (RAG) | PubMed+Guidelines[117]+UpToDate[118]+Dynamed[119] |
| | | QA-RAG[120] | ChatGPT | RAG | FDA QA[120] |
| | | Almanac[121] | ChatGPT | RAG & CoT | Clinical QA[121] |
| | | Oncology-GPT-4[93] | GPT-4 | RAG & ICL | Oncology Guidelines from ASCO and ESMO |

## 2.1 Pre-training

Pre-training typically involves training an LLM on a large corpus of medical texts, including both structured and unstructured text, to learn the rich medical knowledge. The corpus may include EHRs[75], clinical notes[23], and medical literature[56]. In particular, PubMed[50], MIMIC-III clinical notes[57], and PubMed Central (PMC) literature[51], are three widely used medical corpora for medical LLM pre-training. A single corpus or a combination of corpora may be used for pre-training. For example, PubMedBERT[52] and ClinicalBERT are pre-trained on PubMed and MIMIC-III, respectively. In contrast, BlueBERT[62] combines both corpora for pre-training; BioBERT[49] is pre-trained on both PubMed and PMC. The University of Florida (UF) Health EHRs are further introduced in pre-training GatorTron[23] and GatorTronGPT[75]. MEDITRON[94] is pre-trained on Clinical Practice Guidelines (CPGs). The CPGs are used to guide healthcare practitioners and patients in making evidence-based decisions about diagnosis, treatment, and management.

To meet the needs of the medical domain, pre-training medical LLMs typically involve refining the following commonly used training objectives in general LLMs: masked language modeling, next sentence prediction, and next token prediction (Please see Box 1 for an introduction of these three pre-training objectives). For example. BERT-series models (e.g., BioBERT[49], PubMedBERT[52], ClinicalBERT[56], and GatorTron[23]) mainly adopt the masked language modeling and the next sentence prediction for pre-training; GPT-series models (e.g., BioGPT[70], and GatorTronGPT[75]) mainly adopt the next token prediction for pre-training. It is worth mentioning that BERT-like Medical LLMs (e.g., BioBERT[49], PubMedBERT[52], Clinical BERT[56]) are originally derived from the general domain BERT or RoBERTa models. To clarify the differences between different models, in our Table 2, we only show the data source used to further construct medical LLMs. In particular, a more recent work[122] provides a systematic review of existing clinical text datasets for LLMs. After pre-training, medical LLMs can learn rich medical knowledge that can be leveraged to achieve strong performance on different medical tasks.

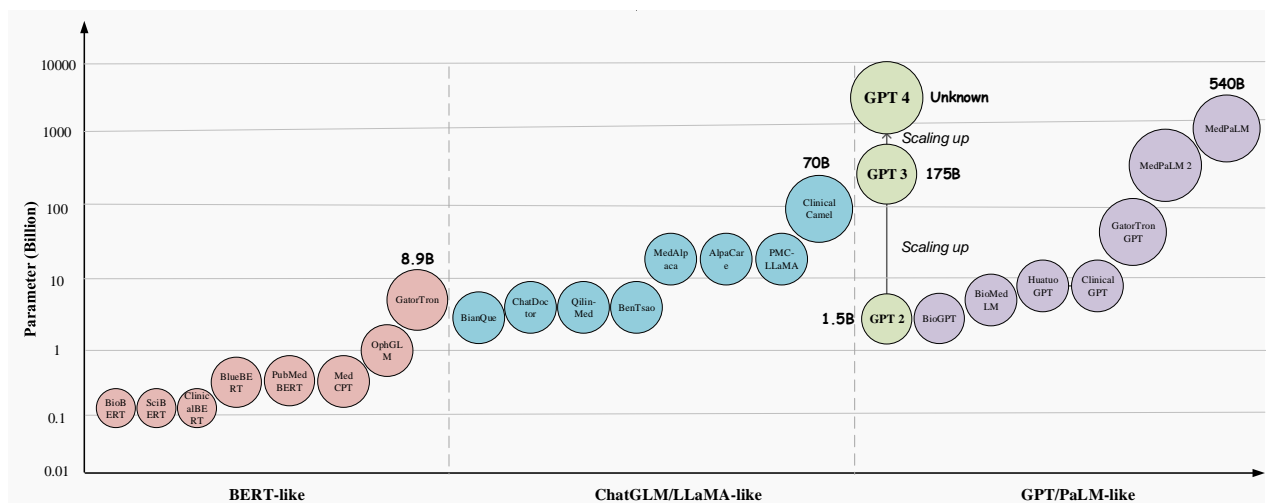

**Figure 2.** We adopt the data from Table 2 to demonstrate the development of model sizes for medical large language models in different model architectures, i.e., BERT-like, ChatGLM/LLaMA-like, and GPT/PaLM-like.

### 2.2 Fine-tuning

It is high-cost and time-consuming to train a medical LLM from scratch, due to its requirement of substantial (e.g., several days or even weeks) computational power and manual labor. One solution is to fine-tune the general LLMs with medical data, and researchers have proposed different fine-tuning methods[11,16,18] for learning domain-specific medical knowledge and obtaining medical LLMs. Current fine-tuning methods include Supervised Fine-Tuning (SFT), Instruction Fine-Tuning (IFT), and Parameter-Efficient Fine-Tuning (PEFT). The resulting fine-tuned medical LLMs are summarized in Table 2. SFT can serve as continued pre-training to fine-tune general LLMs on existing (usually unlabeled) medical corpora. IFT focuses on fine-tuning general LLMs on instruction-based medical data containing additional (usually human-constructed) instructions.

**Supervised Fine-Tuning (SFT)** aims to leverage high-quality medical corpus, which can be physician-patient conversations[15], medical question-answering[16], and knowledge graphs[80,17]. The constructed SFT data serves as a continuation of the pre-training data to further pre-train the general LLMs with the same training objectives, e.g., next token prediction. SFT provides an additional pre-training phase that allows the general LLMs to learn rich medical knowledge and align with the medical domain, thus transforming them into specialized medical LLMs.

The diversity of SFT enables the development of diverse medical LLMs by training on different types of medical corpus. For example, DoctorGLM[76] and ChatDoctor[15] are obtained by fine-tuning the general LLMs ChatGLM[8,9] and LLaMA[4] on the physician-patient dialogue data, respectively. MedAlpaca[16] based on the general LLM Alpaca[35] is fine-tuned using over 160,000 medical QA pairs sourced from diverse medical corpora. Clinicalcamel[18] combines physician-patient conversations, clinical literature, and medical QA pairs to refine the LLaMA-2 model[5]. In particular, Qilin-Med[80] and Zhongjing[89] are obtained by incorporating the knowledge graph to perform fine-tuning on the Baichuan[36] and LLaMA[4], respectively.

In summary, existing studies have demonstrated the efficacy of SFT in adapting general LLMs to the medical domain. They show that SFT improves not only the model's capability for understanding and generating medical text, but also its ability to provide accurate clinical decision support[123].

**Instruction Fine-Tuning (IFT)** constructs instruction-based training datasets[124,123,1], which typically comprise instruction-input-output triples, e.g., instruction-question-answer. The primary goal of IFT is to enhance the model's ability to follow various human/task instructions, align their outputs with the medical domain, and thereby produce a specialized medical LLM.

Thus, the main difference between SFT and IFT is that the former focuses primarily on injecting medical knowledge into a general LLM through continued pre-training, thus improving its ability to understand the medical text and accurately predict the next token. In contrast, IFT aims to improve the model's *instruction following* ability and adjust its outputs to match the

**6/31**

given instructions, rather than accurately predicting the next token as in SFT[124]. As a result, SFT emphasizes the *quantity* of training data, while IFT emphasizes their *quality* and *diversity*. Since IFT and SFT are both capable of improving model performance, there have been some recent works[89,80,88] attempting to combine them for obtaining robust medical LLMs.

In other words, to enhance the performance of LLMs through IFT, it is essential to ensure that the training data for IFT are of high quality and encompass a wide range of medical instructions and medical scenarios. To this end, MedPaLM-2[11] invited qualified medical professionals to develop the instruction data for fine-tuning the general PaLM. BenTsao[17] and ChatGLM-Med[125] constructed the knowledge-based instruction data from the knowledge graph. Zhongjing[89] further incorporated the multi-turn dialogue as the instruction data to perform IFT. MedAlpaca[16] simultaneously incorporated the medical dialogues and medical QA pairs for instruction fine-tuning.

Recent advancements in multimodal LLMs have expanded the capabilities of LLMs to process complex and multimodal medical data. Notable examples include Med-Flamingo[99], LLaVA-Med[102], and Med-Gemini[108]. Med-Flamingo[99] undergoes IFT on medical image-text data, learning to identify abnormalities and generate diagnostic reports. LLaVA-Med's[102] two-stage IFT process involves aligning medical concepts across visual and textual modalities, followed by fine-tuning the model on diverse medical instructions. Med-Gemini's[108] IFT utilizes a curated dataset of medical instructions and multimodal data, enabling it to comprehend complex medical concepts, procedures, and diagnostic reasoning. Meanwhile, MAIRA-1[104] and RadFM[106] are two multimodal LLMs specifically designed for radiology applications. MAIRA-1[104] employs IFT on a dataset of radiology instructions and corresponding medical images, enabling it to analyze radiological images and generate accurate diagnostic reports. RadFM[106], on the other hand, leverages a pre-training approach on a large corpus of radiology-specific image-text data, followed by instruction fine-tuning on a diverse set of radiology instructions. These models' multimodal IFT approaches enable them to bridge the gap between visual and textual medical information, perform a wide range of medical tasks accurately, and generate context-aware responses to complex medical queries.

*Parameter-Efficient Fine-Tuning (PEFT)* aims to substantially reduce computational and memory requirements for fine-tuning general LLMs. The main idea is to keep most of the parameters in pre-trained LLMs unchanged, by fine-tuning only the smallest subset of parameters (or additional parameters) in these LLMs. Commonly used PEFT techniques include Low-Rank Adaptation (LoRA)[126], Prefix Tuning[127], and Adapter Tuning[128,129].

In contrast to fine-tuning full-rank weight matrices, 1) **LoRA** preserves the parameters of the original LLMs and only adds trainable low-rank matrices into the self-attention module of each Transformer layer[126]. Therefore, LoRA can substantially reduce the number of trainable parameters and improve the efficiency of fine-tuning, while still enabling the fine-tuned LLM to capture effectively the characteristics of the tasks. 2) **Prefix Tuning** takes a different approach from LoRA by adding a small set of continuous task-specific vectors (i.e. "prefixes") to the input of each Transformer layer[127,1]. These prefixes serve as the additional context to guide the generation of the model without changing the original pre-trained parameter weights. 3) **Adapter Tuning** involves introducing small neural network modules, known as adapters, into each Transformer layer of the pre-trained LLMs[130]. These adapters are fine-tuned while keeping the original model parameters frozen[130], thus allowing for flexible and efficient fine-tuning. The number of trainable parameters introduced by adapters is relatively small, yet they enable the LLMs to adapt to clinical scenarios or tasks effectively.

In general, PEFT is valuable for developing LLMs that meet unique needs in specific (e.g., medical) domains, due to its ability to reduce computational demands while maintaining the model performance. For example, medical LLMs DoctorGLM[76], MedAlpaca[16], Baize-Healthcare[85], Zhongjing[89], CPLLM[90], and Clinical Camel[18] adopted the LoRA[126] to perform parameter-efficient fine-tuning to efficiently align the general LLMs to the medical domain.

## 2.3 Prompting

Fine-tuning considerably reduces computational costs compared to pre-training, but it requires further model training and collections of high-quality datasets for fine-tuning, thus still consuming some computational resources and manual labor. In contrast, the "prompting" methods efficiently align general LLMs (e.g., PaLM[3]) to the medical domain (e.g., MedPaLM[10]), without training any model parameters. Popular prompting methods include In-Context Learning (ICL), Chain-of-Thought (CoT) prompting, Prompt Tuning, and Retrieval-Augmented Generation (RAG).

**In-Context Learning (ICL)** aims to directly give instructions to prompt the LLM to perform a task efficiently. In general, the ICL consists of four process: task understanding, context learning, knowledge reasoning, and answer generation. First, the model must understand the specific requirements and goals of the task. Second, the model learns to understand the contextual information related to the task with argument context. Then, use the model's internal knowledge and reasoning capabilities to understand the patterns and logic in the example. Finally, the model generates the task-related answers. The advantage of ICL is that it does not require a large amount of labeled data for fine-tuning. Based on the type and number of input examples, ICL can be divided into three categories[131]. 1) **One-shot Prompting**: Only one example and task description are allowed to be entered. 2) **Few-shot Prompting**: Allows the input of multiple instances and task descriptions. 3) **Zero-shot Prompting**: Only task descriptions are allowed to be entered. ICL presents the LLMs making task predictions based on contexts augmented with a few examples and task demonstrations. It allows the LLMs to learn from these examples or demonstrations to accurately perform the task and follow the given examples to give corresponding answers [6]. Therefore, ICL allows LLMs to accurately understand and respond to medical queries. For example, MedPaLM [10] substantially improves the task performance by providing the general LLM, PaLM [3], with a small number of task examples such as medical QA pairs.

**Chain-of-Thought (CoT) Prompting** further improves the accuracy and logic of model output, compared with In-Context Learning. Specifically, through prompting words, CoT aims to prompt the model to generate intermediate steps or paths of reasoning when dealing with downstream (complex) problems [111]. Moreover, CoT can be combined with few-shot prompting by giving reasoning examples, thus enabling medical LLMs to give reasoning processes when generating responses. For tasks involving complex reasoning, such as medical QA, CoT has been shown to effectively improve model performance [10,11]. Medical LLMs, such as DeID-GPT [112], MedPaLM [10], and MedPrompt [12], use CoT prompting to assist them in simulating a diagnostic thought process, thus providing more transparent and interpretable predictions or diagnoses. In particular, MedPrompt [12] directly prompts a general LLM, GPT-4 [7], to outperform the fine-tuned medical LLMs on medical QA without training any model parameters.

**Prompt Tuning** aims to improve the model performance by employing both prompting and fine-tuning techniques [132,129]. The prompt tuning method introduces learnable prompts, i.e. trainable continuous vectors, which can be optimized or adjusted during the fine-tuning process to better adapt to different medical scenarios and tasks. Therefore, they provide a more flexible way of prompting LLMs than the "prompting alone" methods that use discrete and fixed prompts, as described above. In contrast to traditional fine-tuning methods that train all model parameters, prompt tuning only tunes a very small set of parameters associated with the prompts themselves, instead of extensively training the model parameters. Thus, prompt tuning effectively and accurately responds to medical problems [12], with minimal incurring computational cost.

Existing medical LLMs that employ the prompting techniques are listed in Table 2. Recently, MedPaLM [10] and MedPaLM-2 [11] propose to combine all the above prompting methods, resulting in Instruction Prompt Tuning, to achieve strong performances on various medical question-answering datasets. In particular, using the MedQA dataset for the US Medical Licensing Examination (USMLE), MedPaLM-2 [11] achieves a competitive overall accuracy of 86.5% compared to human experts (87.0%), surpassing previous state-of-the-art method MedPaLM [10] by a large margin (19%).

**Retrieval-Augmented Generation (RAG)** enhances the performance of LLMs by integrating external knowledge into the generation process. In detail, RAG can be used to minimize LLM's hallucinations, obscure reasoning processes, and reliance on outdated information by incorporating external database knowledge [133]. RAG consists of three main components: retrieval, augmentation, and generation. The retrieval component employs various indexing strategies and input query processing techniques to search and top-ranked relevant information from an external knowledge base. The retrieved external data is then augmented into the LLM's prompt, providing additional context and grounding for the generated response. By directly updating the external knowledge base, RAG mitigates the risk of catastrophic forgetting associated with model weight modifications, making it particularly suitable for domains with low error tolerance and rapidly evolving information, such as the medical field. In contrast to traditional fine-tuning methods, RAG enables the timely incorporation of new medical information without compromising the model's previously acquired knowledge, ensuring the generated outputs remain accurate and up-to-date in the face of evolving medical challenges. Most recently, researchers proposed the first benchmark MIRAGE [134] based on medical information RAG, including 7,663 questions from five medical QA datasets, which has been established to both steer research and facilitate the practical deployment of medical RAG systems

In RAG, retrieval can be achieved by calculating the similarity between the embeddings of the question and document chunks, where the semantic representation capability of embedding models plays a key role. Recent research has introduced prominent embedding models such as AngIE [135], Voyage [136], and BGE [137]. In addition to embedding, the retrieval process can be optimized via various strategies such as adaptive retrieval, recursive retrieval, and iterative retrieval [138,139,140]. Several recent works have demonstrated the effectiveness of RAG in medical and pharmaceutical domains. Almanac [121] is a large language framework augmented with retrieval capabilities for medical guidelines and treatment recommendations, surpassing the performance of ChatGPT on clinical scenario evaluations, particularly in terms of completeness and safety. Another work QA-RAG [120] employs RAG with LLM for pharmaceutical regulatory tasks, where the model searches for relevant guideline documents and provides answers based on the retrieved guidelines. Chat-Orthopedist [116], a retrieval-augmented LLM, assists adolescent idiopathic scoliosis (AIS) patients and their families in preparing for meaningful discussions with clinicians by providing accurate and comprehensible responses to patient inquiries, leveraging AIS domain knowledge.

## 2.4 Discussion

This section discusses the principles of medical LLMs, including three types of methods for building models: pre-training, fine-tuning, and prompting. To meet the needs of practical medical applications, users can choose proper medical LLMs according to the magnitude of their own computing resources. Companies or institutes with massive computing resources can either pre-train an application-level medical LLM from scratch or fine-tune existing open-source general LLM models (e.g., LLaMA[43]) using large-scale medical data. The results in existing literature (e.g., MedPaLM-2[11], MedAlpaca[16] and Clinical Camel[18]) have shown that fine-tuning general LLMs on medical data[122] can boost their performance of medical tasks. For example, Clinical Camel[18], which is fine-tuned on the LLaMA-2-70B[5] model, even outperforms GPT-4[18]. However, for small enterprises or individuals with certain computing resources, combining with the understanding of medical tasks and a reasonable combination of ICL, prompting engineering, and RAG, to prompt black-box LLMs may also achieve strong results. For example, MedPrompt[12] stimulates the commercial LLM GPT-4[7] through an appropriate combination of prompt strategies to achieve comparable or even better results than fine-tuned medical LLMs (e.g., MedPaLM-2[11]) and human experts, suggesting that a mix of prompting strategies is an efficient and green solution in the medical domain rather than fine-tuning.

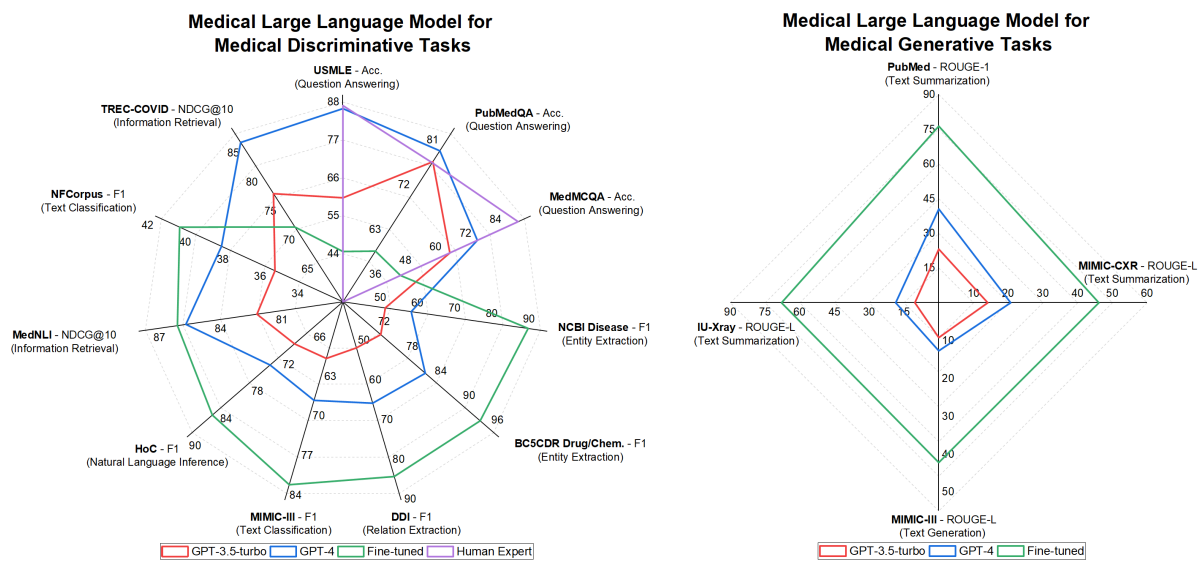

**Figure 3.** Performance (Dataset-Metric (Task)) comparison between the GPT-3.5 turbo, GPT-4, state-of-the-art task-specific lightweight models (Fine-tuned), and human experts, on seven medical tasks across eleven datasets. All data presented in our Figures originates from published and peer-reviewed literature. Please refer to the supplementary material for the detailed data.

## 3 Medical Tasks

In this section, we will introduce two popular types of medical machine learning tasks: generative and discriminative tasks, including ten representative tasks that further build up clinical applications. Figure 3 illustrates the performance comparisons between different LLMs. For clarity, we will only cover a general discussion of those tasks. The detailed definition of the task and the performance comparisons can be found in our supplementary material.

### 3.1 Discriminative Tasks

Discriminative tasks are for categorizing or differentiating data into specific classes or categories based on given input data. They involve making distinctions between different types of data, often to categorize, classify, or extract relevant information from structured text or unstructured text. The representative tasks include Question Answering, Entity Extraction, Relation Extraction, Text Classification, Natural Language Inference, Semantic Textual Similarity, and Information Retrieval.

The typical input for discriminative tasks can be medical questions, clinical notes, medical documents, research papers, and patient EHRs. The output can be labels, categories, extracted entities, relationships, or answers to specific questions, which are often structured and categorized information derived from the input text. In existing LLMs, the discriminative tasks are widely studied and used to make predictions and extract information from input text.

For example, based on medical knowledge, medical literature, or patient EHRs, the medical question answering (QA) task can provide precise answers to clinical questions, such as symptoms, treatment options, and drug interactions. This can help clinicians make more efficient and accurate diagnoses[10,11,19]. Entity extraction can automatically identify and categorize critical information (i.e. entities) such as symptoms, medications, diseases, diagnoses, and lab results from patient EHRs, thus assisting in organizing and managing patient data. The following entity linking task aims to link the identified entities in a structured knowledge base or a standardized terminology system, e.g., SNOMED CT[141], UMLS[115], or ICD codes[142]. This task is critical in clinical decision support or management systems, for better diagnosis, treatment planning, and patient care.

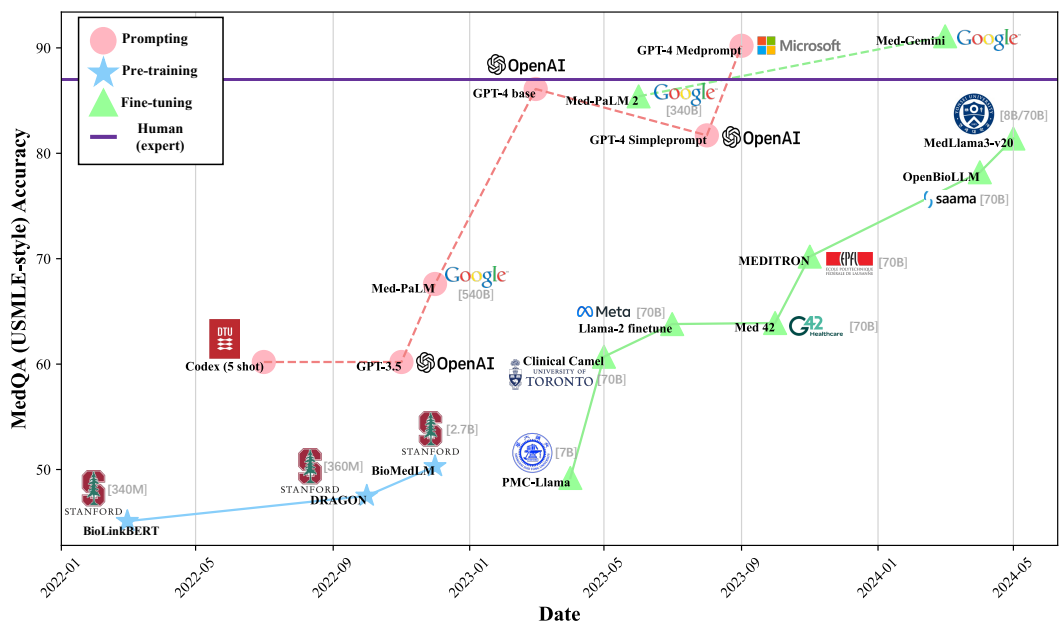

**Figure 4.** We demonstrate the development of medical large language models over time in different model development types through the scores of the United States Medical Licensing Examination (USMLE) from the MedQA dataset. Solid and dashed lines represent open-source and closed-source models, respectively.

### 3.2 Generative Tasks

Different from discriminative tasks that focus on understanding and categorizing the input text, generative tasks require a model to accurately generate fluent and appropriate *new* text based on given inputs. These tasks include medical text summarization[143,144], medical text generation[70], and medical text simplification[145].

For medical text summarization, the input and output are typically long and detailed medical text (e.g., "Findings" in radiology reports), and a concise summarized text (e.g., the "Impression" in radiology reports). Such text contains important medical information that enables clinicians and patients to efficiently capture the key points without going through the entire text. It can also help medical professionals to draft clinical notes by summarizing patient information or medical histories.

In medical text generation, e.g., discharge instruction generation[146], the input can be medical conditions, symptoms, patient demographics, or even a set of medical notes or test results. The output can be a diagnosis recommendation of a medical condition, personalized instructional information, or health advice for the patient to manage their condition outside the hospital.

Medical text simplification[145] aims to generate a simplified version of the complex medical text by, for example, clarifying and explaining medical terms. Different from text summarization, which concentrates on giving shortened text while maintaining most of the original text meanings, text simplification focuses more on the readability part. In particular, complicated or opaque words will be replaced; complex syntactic structures will be improved; and rare concepts will be explained[38]. Thus, one example application is to generate easy-to-understand educational materials for patients from complex medical texts. It is useful for making medical information accessible to a general audience, without altering the essential meaning of the texts.

### 3.3 Performance Comparisons

Figure 3 shows that some existing general LLMs (e.g., GPT-3.5-turbo and GPT-4[7]) have achieved strong performance on existing medical machine learning tasks. This is most noticeable for the QA task where GPT-4 (shown in the blue line in Figure 3) consistently outperforms existing task-specific fine-tuned models and is even comparable to human experts (shown in the purple line). The QA datasets of evaluation include MedQA (USMLE)[14], PubMedQA[147], and MedMCQA[148]. To better understand the QA performance of existing medical LLMs, in Figure 4, we further demonstrate the QA performance of medical LLMs on the MedQA dataset over time in different model development types. It also clearly shows that current works, e.g., MedPrompt[12] and Med-Gemini[107,108], have successfully proposed several prompting and fine-tuning methods to enable LLMs to outperform human experts.

However, on the non-QA discriminative tasks and generative tasks, as shown in Figure 3, the existing general LLMs perform worse than the task-specific fine-tuned models. For example, for the non-QA discriminative tasks, the state-of-the-art task-specific fine-tuned model BioBERT[49] achieves an F1 score of 89.36, substantially exceeding the F1 score of 56.73 by

**10/31**

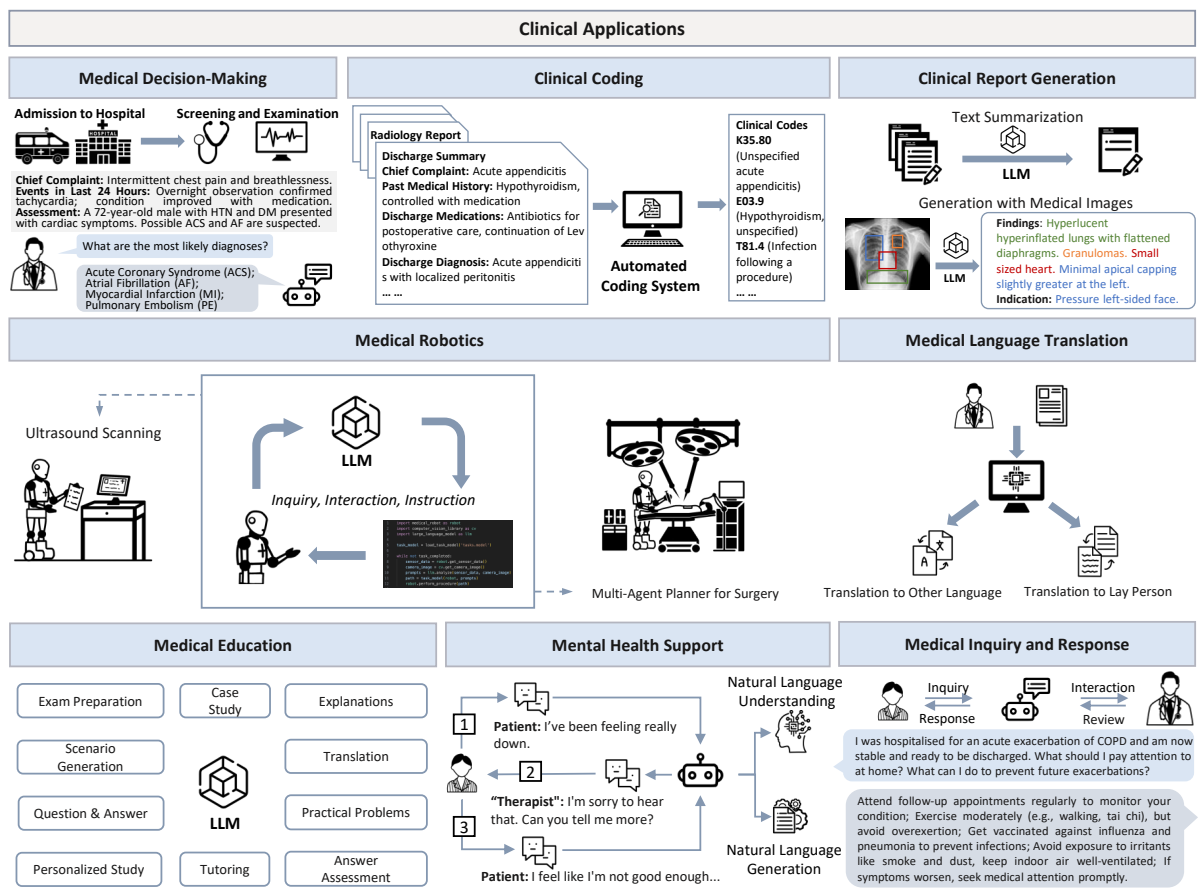

**Figure 5.** Integrated overview of potential applications [114,150,151,152,153] of large language models in medicine.

GPT-4, on the entity extraction task using the NCBI disease dataset [149]. For the generative tasks, we can see that the strong LLM GPT-4 clearly underperforms task-specific lightweight models on all datasets. We hypothesize that the reason for the strong QA capability of the current general LLMs is that the QA task is close-ended; i.e. the correct answer is already provided by multiple candidates. In contrast, most non-QA tasks are open-ended where the model has to predict the correct answer from a large pool of possible candidates, or even without any candidates provided.

Overall, the comparison proves that the current general LLMs have a strong question-answering capability, however, the capability on other tasks still needs to be improved. In detail, current LLMs are comparable to state-of-the-art models and human experts on the exam-style close-ended QA task with provided answer options. However, real-world open clinical practice usually involves answering open-ended questions without pre-set options and diverges far from the structured nature of exam-taking. The poor performance of LLMs in other non-QA tasks without options indicates a considerable need for advancement before LLMs can be integrated into the actual clinical decision-making process without answer options. Therefore, we advocate that the evaluation of medical LLMs should be extended to a broad range of tasks including non-QA tasks, instead of being limited mainly to medical QA tasks. Hereafter, we will discuss specific clinical applications of LLMs, followed by their challenges and future directions.

## 4 Clinical Applications

Currently, most of existing medical LLMs are still in the research and development stage, with limited application and validation in real-world clinical scenarios. However, some initial attempts and explorations have begun to emerge. Researchers are also exploring the integration of large language models into clinical decision support systems to provide evidence-based recommendations and references [155,55,154]. Additionally, some research teams are developing tools based on large language models to assist in clinical trial recruitment by analyzing electronic health records to identify eligible participants [156]. Some healthcare institutions are experimenting with using LLMs for clinical coding and formatting to improve efficiency and accuracy in medical billing and reimbursement [161,162,163]. Researchers are investigating the use of LLMs for clinical report generation, aiming to automate the process of creating coherent and accurate medical reports based on patient data[113,104,106]. LLMs are being integrated into medical robotics to enhance decision-making, collaboration, and diagnostic capabilities, improving surgical precision and efficiency[167,169,170]. In the realm of medical language translation, efforts are being made to utilize LLMs to translate medical information into multiple languages for foreign patients[171,174,175] and to simplify complex medical terminology for lay people[176,177], enhancing patient understanding and communication. In the field of medical education, LLMs are being considered as tools to enhance learning experiences by providing personalized content, answering questions, and offering interactive educational materials[178,108]. Certain organizations are testing the use of chatbots or virtual assistants to provide mental health support, aiming to increase the accessibility of mental health services[179,180,181]. Furthermore, researchers are developing LLM-based systems for medical inquiry and response, aiming to provide accurate and timely answers to patients' questions, triage inquiries, and assist healthcare professionals in addressing common concerns[189,190,191].

**Table 3.** Summary of existing medical LLMs tailored to various clinical applications, in terms of their architecture, model development, the number of parameters, the scale of PT/FT data, and the data source. M: million, B: billion. PT: pre-training. FT: fine-tuning. ICL: in-context learning. CoT: chain-of-thought prompting. RAG: retrieval-augmented generation. This information provides guidelines on how to select and build models. We further provide the evaluation information (i.e., task and performance) to show how current works evaluate their models.

| Application | Model | Architecture | Model Development | # Params | Data Scale | Data Source | Evaluation (Task: Score) |
|---|---|---|---|---|---|---|---|
| Medical Decision-Making (Sec. 4) | Dr. Knows[114] | GPT-3.5 | ICL | 154B | 5820 notes | MIMIC-III[57]+IN-HOUSE[114] | Diagnosis Summarization: 30.72 ROUGE-L |
| | DDx PaLM-2[154] | PaLM-2 | FT & ICL | 340B | - | MultiMedQA[11]+MIMIC-III[57] | Differential Diagnosis: 0.591 top-10 Accuracy |
| | NYUTron[55] | BERT | PT & FT | 110M | 7.25M notes, 4.1B tokens | NYU Notes[55] | Readmission Prediction: 0.799 AUC<br>In-hospital Mortality Prediction: 0.949 AUC<br>Comorbidity Index Prediction: 0.894 AUC<br>Length of Stay Prediction: 0.787 AUC<br>Insurance Denial Prediction: 0.872 AUC |
| | Foresight[155] | GPT-2 | PT & FT | 1.5B | 35M notes | King's College Hospital, MIMIC-III South London and Maudsley Hospital | Next Biomedical Concept Forecast: 0.913 F1 |
| | TrialGPT[156] | GPT-4 | - | - | 184 patients | 2016 SIGIR[157], 2021 & 2022 TREC[158] | Ranking Clinical Trials: 0.733 P@10, 0.817 NDCG@10<br>Excluding clinical trials: 0.775 AUROC |
| Clinical Coding (Sec. 4) | PLM-ICD[159] | RoBERTa | FT | 355M | 70,539 notes | MIMIC-II[160]+MIMIC-III[57] | ICD Code Prediction: 0.926 AUC, 0.104 F1 |
| | DRG-LLaMA[161] | LLaMA-7B | FT | 7B | 25k pairs | MIMIC-IV[105] | Diagnosis-related Group Prediction: 0.327 F1 |
| | ChatICD[162] | ChatGPT | ICL | - | 10k pairs | MIMIC-III[57] | ICD Code Prediction: 0.920 AUC, 0.681 F1 |
| | LLM-codex[163] | ChatGPT+LSTM | ICL | - | - | MIMIC-III[57] | ICD Code Prediction: 0.834 AUC, 0.468 F1 |
| Clinical Report Generation (Sec. 4.3) | ImpressionGPT[164] | ChatGPT | ICL & RAG | 110M | 184k reports | MIMIC-CXR[105]+IU X-ray[165] | Report Summarization: 47.93 ROUGE-L |
| | RadAdapt[166] | T5 | FT | 223M, 738M | 80k reports | MIMIC-III[57] | Report Summarization: 36.8 ROUGE-L |
| | ChatCAD[113] | GPT-3 | ICL | 175B | 300 reports | MIMIC-CXR[105] | Report Generation: 0.605 F1 |
| | MAIRA-1[104] | ViT+Vicuna-7B | FT | 8B | 337k pairs | MIMIC-CXR[105] | Report Generation: 28.9 ROUGE-L |
| | RadFM[106] | ViT+LLaMA-13B | PT & FT | 14B | 32M pairs | MedMD[106] | Report Generation: 18.22 ROUGE-L |
| Medical Robotics (Sec. 4.4) | SuFIA[167] | GPT-4 | ICL | - | 4 tasks | ORBIT-Surgical[168] | Surgical Tasks: 100 Success Rate |
| | UltrasoundGPT[169] | GPT-4 | ICL | - | 522 tasks | - | Task Completion: 80 Success Rate |
| | Robotic X-ray[170] | GPT-4 | ICL | - | - | - | X-ray Surgery: 7.6/10 Human Rating |
| Medical Language Translation (Sec. 4.5) | Medical mT5[171] | T5 | PT | 738M, 3B | 4.5B pairs | PubMed[50]+EMEA[172] ClinicalTrials[173], etc. | (Multi-Task) Sequence Labeling: 0.767 F1<br>Augment Mining 0.733 F1 |
| | Apollo[174] | Qwen | PT & FT | 1.8B-7B | 2.5B pairs | ApolloCorpora[174] | QA: 0.588 Accuracy |
| | BiMediX[175] | Mistral | FT | 13B | 1.3M pairs | BiMed1.3M[175] | Question Answering: 0.654 Accuracy |
| | Biomed-sum[176] | BART | FT | 406M | 27k papers | BioCiteDB[176] | Abstractive Summarization: 32.33 ROUGE-L |
| | RALL[177] | BART | FT & RAG | 406M | 63k pairs | CELLS[176] | Lay Language Generation: N/A |
| Medical Education (Sec. 4.6) | ChatGPT[178] | GPT-3.5/GPT-4 | ICL | - | - | - | Curriculum Generation, Learning Planning |
| | Med-Gemini[108] | Gemini | FT & CoT | - | - | MedQA-R/RS[108]+MultiMedQA[11] MIMIC-III[57]+MultiMedBench[109] | Text-based QA: 0.911 Accuracy<br>Multimodal QA: 0.935 Accuracy |
| Mental Health Support (Sec. 4.7) | PsyChat[179] | ChatGLM | FT | 6B | 350k pairs | Xingling[179]+Smilechat[179] | Text Generation: 27.6 ROUGE-L |
| | ChatCounselor[180] | Vicuna | FT | 7B | 8k instructions | Psych8K[180] | Question Answering: Evaluated by ChatGPT |
| | Mental-LLM[181] | Alpaca, FLAN-T5 | FT & ICL | 7B, 11B | 31k pairs | Dreaddit[182]+DepSeverity[183]+SDCNL[184] CSSRS-Suicide[185]+Red-Sam[186] Twt-60Users[187]+SAD[188] | Mental Health Prediction: 0.741 Accuracy |
| Medical Inquiry and Response (Sec. 4.8) | AMIE[189] | PaLM2 | FT | 340B | >2M pairs | MedQA[14]+MultiMedBench[109] MIMIC-III[57]+real-world diaglogue[189] | Diagnostic Accuracy: 0.920 Top-10 Accuracy |
| | Healthcare Copilot[190] | ChatGPT | ICL | - | - | MedDialog[190] | Inquiry Capability: 4.62/5 (ChatGPT)<br>Conversational Fluency: 4.06/5 (ChatGPT)<br>Response Accuracy: 4.56/5 (ChatGPT)<br>Response Safety: 3.88/5 (ChatGPT) |
| | Conversational Diagnosis[191] | GPT-4/LLaMA | ICL | - | 40k pairs | MIMIC-IV[92] | Disease Screening: 0.770 Top-10 Hit Rate<br>Differential Diagnosis: 0.910 Accuracy |

To this end, as shown in Figure 5, we will introduce the clinical applications of LLMs in this section. Each subsection contains a specific application and discusses how LLMs perform this task. Table 3 summarizes the guidelines on how to select, build, and evaluate medical LLMs for various clinical applications. Although there are currently no large-scale clinical trials specifically targeting these models, researchers are actively evaluating their effectiveness and safety in various healthcare settings. As research progresses and evidence accumulates, it is expected that the application of these large language models in

## 4.1 Medical Decision-Making

Medical decision-making, including diagnosis, prognosis, treatment suggestion, risk prediction, clinical trial matching, etc., heavily relies on the synthesis and interpretation of vast amounts of information from various sources, such as patient medical histories, clinical data, and the latest medical literature. The advent of LLMs has opened up new opportunities for enhancing these critical processes in healthcare. These advanced models can rapidly process and comprehend massive volumes of medical data, literature, and legal guidelines, potentially aiding healthcare professionals in making more informed and legally sound decisions across a wide range of clinical scenarios[192,19]. For instance, in medical diagnosis, LLMs can assist practitioners in analyzing objective medical data from tests and self-described subjective symptoms to conclude the most likely health problem occurring in the patient[192]. Similarly, LLMs can support treatment planning by providing personalized recommendations based on the latest clinical evidence and patient-specific factors[19]. Furthermore, LLMs can contribute to prognosis and risk prediction by identifying patterns and risk factors from large-scale patient data, enabling more accurate and timely interventions[55]. By leveraging the power of LLMs, healthcare professionals can enhance their decision-making capabilities across the spectrum of clinical tasks, leading to improved patient outcomes and more efficient healthcare delivery.

**Guideline** To create an effective LLM-based medical decision-making system, practitioners should begin with a robust LLM and enhance it with specialized medical knowledge. This section outlines key strategies and examples of successful implementations in this field. Dr. Knows[114] demonstrates the efficacy of integrating knowledge graphs from the Unified Medical Language System (UMLS)[115] to improve diagnosis prediction and provide treatment suggestions. This approach involves fine-tuning T5 models[193] with extracted diagnoses as prompts and employing zero-shot prompting for LLMs like ChatGPT. Alternatively, models like DDx PaLM-2[154] showcase the potential of fine-tuning LLMs (such as Google's PaLM-2) with extensive medical datasets. This approach enables interactive diagnosis assistance, supporting clinicians in identifying potential diagnoses for better medical decision-making. NYUTron[55] is pretrained and fine-tuned on various NYU hospitals and is capable of three clinical tasks (in-patient mortality prediction, comorbidity index prediction, and readmission prediction) and two operational tasks (insurance claim denial prediction and inpatient LOS prediction). Foresight[155], is another model which trained on UK hospital patient data and can be used for forecasting the risk of disorders, differential diagnoses, suggest substances (e.g., to do with medicines, allergies, or poisonings) to be used, etc. For clinical trial matching, TrialGPT[156] presents a novel GPT-4-based framework that accurately predicts criterion-level eligibility with faithful explanations, reducing screening time for human experts. Evaluating LLM-based medical diagnosis systems requires task-specific approaches. For general diagnostic accuracy, metrics like AUC, precision, recall, and F1 score are used with annotated datasets[154,155,156]. Some works evaluate the diagnostics with free-text using ROUGE score and CUI F-score[114]. Crucially, all evaluations must include expert clinician review to ensure clinical relevance and potential real-world impact.

**Discussion** One distinct limitation of using LLMs as the sole tool for medical diagnosis is the heavy reliance on subjective text inputs from the patient. Since LLMs are text-based, they lack the inherent capability to analyze medical diagnostic imagery. Given that objective medical diagnoses frequently depend on visual images, LLMs are often unable to directly conduct diagnostic assessments as they lack concrete visual evidence to support disease diagnosis[194]. However, they can help with diagnosis as a logical reasoning tool for improving accuracy in other vision-based models. One such example is ChatCAD[113], where images are first fed into an existing computer-aided diagnosis (CAD) model to obtain tensor outputs. These outputs are translated into natural language, which is subsequently fed into ChatCAD to summarize results and formulate diagnoses. ChatCAD achieves a recall score of 0.781, substantially higher than that (0.382) of the state-of-the-art task-specific model. Nevertheless, all the aforementioned methods of implementing LLMs cannot directly process images; instead, they either rely on transforming images into text beforehand or rely on an external separate vision encoder to embed images.

## 4.2 Clinical Coding

Clinical coding, such as the International Classification of Diseases (ICD)[142], medication coding, and procedure coding, plays a crucial role in healthcare by standardizing diagnostic, procedural, and treatment information. These codes are essential for tracking health metrics, treatment outcomes, billing, and reimbursement processes. However, the manual entry of these codes is time-consuming and prone to errors. Large language models (LLMs) have shown promise in automating the clinical coding process by extracting relevant medical terms from clinical notes and assigning corresponding codes, including ICD codes[159,161,162,163], medication codes (e.g., National Drug Code[195] or RxNorm[196]), and procedure codes (e.g., Current Procedural Terminology[197]). By leveraging the vast medical knowledge and natural language understanding capabilities of LLMs, healthcare professionals can benefit from reduced workload and improved accuracy in clinical coding.

**Guideline** When developing LLM-based applications for ICD coding, several notable examples can serve as guidance and inspiration, especially in automating the process of ICD coding. PLM-ICD[159] is an example of an LLM-based approach that builds upon the RoBERTa model[39], fine-tuning it specifically for ICD coding. It utilizes a domain-specific base model with medicine-specific knowledge to enhance its ability to understand medical terms and achieves strong performance on 70,539 notes from the MIMIC-II and MIMIC-III datasets[57]. Other LLM-based approaches for ICD coding include DRG-LLaMA[161], which leverages the LLaMA model and applies parameter-efficient fine-tuning techniques, such as LoRA, to adapt the model to this task. ChatICD[162] and LLM-codex[163] both utilize the ChatGPT model with prompts for ICD coding, with LLM-codex[163] taking a step further by training an LSTM model on top of the ChatGPT responses, demonstrating its strong performance.

ICD coding is typically formulated as a multi-label classification task, with most work in this area utilizing the MIMIC-III dataset for training and evaluation. Models are assessed based on their F1 score, AUC, and Precision@k, considering either the top k most frequent labels or the full label set. The development of LLMs for ICD coding has the potential to reduce the manual effort required from healthcare professionals, improve the accuracy and consistency of coded data, and facilitate more efficient billing and reimbursement processes.

**Discussion**   One challenge while deploying LLMs for clinical coding is the potential biases and hallucinations. In particular, traditional multi-label classification models can easily constrain their outputs to a predefined list of (usually >1000) candidate codes through a classification neural network. In contrast, generative LLMs can suffer from major hallucinations while the input text is lengthy. As a result, the LLM may assign an code that is not in the candidate list or a non-existent clinical code to the input text. It leads to confusion when interpreting medical records[23] and is, therefore, crucial to establish a proactive mechanism to detect and rectify errors before they enter patient EHRs.

Currently, the majority of research on LLMs for clinical coding focuses on ICD coding due to its widespread use and the availability of large-scale datasets, such as MIMIC-III, which provide ample training data for model development and evaluation. However, there is a growing need for LLMs that can be applied to other types of clinical coding, such as medication and procedure coding. These coding systems are equally important for accurately capturing patient information, facilitating billing and reimbursement processes, and supporting clinical decision-making. Expanding the capabilities of LLMs to encompass medication and procedure coding would greatly enhance the efficiency and accuracy of the clinical coding process. By leveraging the vast knowledge and natural language understanding capabilities of LLMs, healthcare professionals could benefit from automated coding systems that accurately extract medication and procedure information from clinical notes, reducing the time and effort required for manual coding.

### 4.3 Clinical Report Generation

Clinical reports, such as radiology reports, discharge summaries, and patient clinic letters, refer to standardized documentation that healthcare workers complete after each patient visit[198]. Therefore, clinical report generation usually involves text generation/summarization, and information retrieval. A large portion of the report is often medical diagnostic results. It is typically tedious for overworked clinicians to write clinical reports, and thus they are often incomplete or error-prone. Meanwhile, LLMs can be used intuitively as a summarization tool to help with clinical report generation. In this instance, LLMs act as an assistant tool for clinicians which helps improve efficiency and reduce potential errors in lengthy reports[164,166].

Another popular approach to generating clinical reports using LLMs involves incorporating a vision-based model to provide complementary information[113,106,104]. The vision model analyzes the input medical image and generates an annotation, which serves as a direct and supplementary input to the LLM alongside additional text prompts. By leveraging the combination of visual and textual information, the LLM produces accurate and fluent reports that adhere to the specified parameters and structure.

**Guideline**   When developing LLM-based applications for radiology report generation, several models can serve as guidance and inspiration, which have different focus. General medical vision-language models like Med-Gemini[108], LlaVA-Med[102], and Med-Flamingo[99] can be serviced as foundation models for the broad medical domain including, radiology, pathology, etc., where there are also models trained specifically on radiographs, such as ChatCAD[113], MAIRA-1[104], and RadFM[106], have shown superior performance in specific subdomains. These models leverage the power of large language models and fine-tune them on domain-specific data to generate radiology reports that accurately capture the relevant information and findings.

An alternative approach to radiology report generation focuses on language models that leverage textual data for report summarization. This can be achieved using either unimodal LLMs, which input a long report and generate a summary, or multimodal LLMs, which input both the long report and the related image to generate a summary. The vision-language models mentioned above can also be developed for report summarization. In terms of unimodal LLMs, ImpressionGPT[164] serves as an example, employing dynamic prompt generation and iterative optimization to generate concise and informative report summaries. RadAdapt[166] systematically evaluates various language models and lightweight adaptation methods, achieving optimal performance through pre-training on clinical text and parameter-efficient fine-tuning with LoRA, while also investigating the impact of few-shot prompting.

When evaluating the performance of LLM-based radiology report generation models, most work relies on the MIMIC-III or MIMIC-IV datasets for training and evaluation, as they are the largest publicly available free-text electronic health records (EHRs). Common automatic evaluation metrics include lexical methods such as BLEU[199], ROUGE[200], and METEOR[201], as well as semantic-based methods like BERTScore[202]. Additionally, radiology-specific metrics such as CheXbert similarity[203], RadGraph[204], and RadCliQ[205] have been developed to better assess the quality and accuracy of the generated reports in the context of radiology.

By leveraging these existing models and evaluation metrics, researchers and developers can create LLM-based radiology report generation applications that accurately and efficiently produce high-quality reports, ultimately improving the efficiency and effectiveness of radiology workflows.

**Discussion** While LLMs have demonstrated the ability to generate clinical reports that are more comprehensive and precise than those written by human counterparts[144], they still face challenges in terms of hallucinations and literal interpretation of inputs, lacking the assumption-based perspective often employed by human doctors. Moreover, LLM-generated reports tend to be less concise compared to human-written ones. The evaluation of LLMs in this domain is particularly challenging due to the specialized nature of the content and the generative nature of the task. Current automatic evaluation methods for clinical report generation primarily focus on lexical metrics, which can lead to biased and inaccurate assessments of the contextual information present in the reports[206]. For instance, consider two sentences with similar meanings but different wordings: "The patient's blood glucose level is within normal limits" and "The patient does not exhibit signs of hyperglycemia". While both convey the absence of hyperglycemia, lexical evaluation metrics may struggle to accurately capture their semantic equivalence, as they rely on direct word-level comparisons. This discrepancy highlights the need for more sophisticated evaluation techniques that can account for the nuances and variations in expressing clinical information. Developing evaluation methods that go beyond surface-level similarities and consider the underlying medical context is crucial for ensuring the reliability and usefulness of LLMs in generating clinical reports.

### 4.4 Medical Robotics

Medical robotics is revolutionizing healthcare, offering precision in various aspects, such as surgical procedures and medical imaging[207]. Recent advancements in incorporating LLMs into medical robotics have shown promising results in enhancing the capabilities of these systems[208]. LLMs serve as a complementary technology to robotics, augmenting their decision-making, communication, interaction, and control abilities. For example, surgical robots assisted with LLMs enable minimally invasive procedures with increased accuracy and reduced patient recovery times[169,208,209]. Multi-agent planning systems designed with LLMs involve the coordination of multiple robotic units to perform collaborative tasks, enhancing surgical accuracy and efficiency[209]. Additionally, in the field of ultrasound and radiology diagnostics, LLMs have been combined with domain knowledge to enable precise diagnostics and dynamic scanning strategies, improving the efficiency and quality of scans[169,170].

**Guideline** Integrating LLMs into medical robotics poses challenges due to healthcare complexities and real-world evaluation difficulties. Nevertheless, three innovative systems from current research exemplify the potential of LLMs in enhancing medical robotics, serving as representative examples in this emerging field. SuFIA[167] showcases the integration of LLMs in robotic surgery. This system combines the advanced reasoning capabilities of LLMs, specifically GPT-4 Turbo, with perception modules to implement high-level planning and low-level control of surgical robots for sub-task execution. In the field of medical imaging, UltrasoundGPT[169] presents an innovative approach to ultrasound-guided procedures. This system equips ultrasound robots with LLMs and domain-specific knowledge, utilizing an ultrasound operation knowledge database to enable precise motion planning. UltrasoundGPT employs a dynamic scanning strategy based on prompt engineering, allowing LLMs to adjust motion planning during procedures. This system demonstrates improved ultrasound scan efficiency and quality through verbal command interpretation, contributing to advancements in non-invasive diagnostics and streamlined workflows. Another noteworthy application involves the interpretation of domain-specific language in X-ray-guided surgery[170]. This work introduces a minimal protocol enabling an LLM, specifically GPT-4, to control a robotic X-ray system, namely the Brainlab Loop-X device. This development showcases the potential of LLMs to enhance the precision and efficiency of X-ray-guided surgical procedures through improved communication between surgeons and imaging systems.

Evaluating such systems clinically can be complicated. The complexity of medical procedures, ethical considerations, and patient safety concerns make it difficult to conduct comprehensive evaluations in actual healthcare environments. Consequently, most current evaluations rely heavily on simulated data and controlled laboratory settings. For instance, SuFIA and Robotic X-ray's performance are assessed using a combination of simulated surgical scenarios and expert human evaluation[167,170]. Similarly, UltrasoundGPT is tested through the assessment of task completion[169].

**Discussion** Integrating LLMs into medical robotics algorithms for route planning and motion control poses a critical challenge due to the risk of errors and biases inherent in LLMs. The complex and dynamic nature of shared human-robot workspaces may lead to LLM-powered medical robots misjudging human intentions or making inappropriate decisions, posing safety risks. Future research opportunities could explore safety features for medical robots, such as sophisticated sensing technologies and physical design constraints, which aim to minimize the occurrence and consequences of judgment errors related to LLMs in shared human-robot environments[210,211,212].

### 4.5 Medical Language Translation

There are two main areas of medical language translation; the translation of medical terminology from one language to another[171,174,175] and the translation of medical dialogue for ease of interpretation by non-professional personnel[176,177]. Both areas are important for seamless communication between different groups. It promotes accurate diagnosis, treatment planning, and medication administration, minimizing medical errors and improving patient safety. By bridging the communication gap between healthcare providers and patients, it fosters informed decision-making, shared understanding, and enhanced patient satisfaction. Moreover, it empowers non-medical personnel to actively participate in patient care, promoting patient-centered care and cultural sensitivity. Effective medical language translation is essential for providing high-quality healthcare to diverse patient populations.

**Guideline** In the development of multilingual LLMs for medical language translation, fine-tuning pre-trained models on parallel corpora of medical texts has proven to be an effective approach. By leveraging diverse datasets such as scientific articles, clinical notes, and medical glossaries, these models can capture the nuances and domain-specific meanings of medical terms across languages. Multilingual LLMs like Medical mT5[171], Apollo[174] and BiMediX[175], which are trained on extensive medical datasets in multiple languages, can be further fine-tuned to accurately translate medical terminology between languages such as English, French, Spanish, Chinese, and Arabic. This enables seamless communication and knowledge sharing among healthcare professionals across linguistic boundaries.

When translating medical dialogue for non-professional understanding, it is crucial to fine-tune LLMs on datasets that encompass both technical medical conversations and their corresponding lay-language explanations. This training approach allows the models to learn the mapping between complex medical jargon and more accessible language, facilitating better comprehension by patients and the general public. Techniques such as retrieval augmentation, which involves retrieving relevant lay-language explanations from external knowledge sources, can further enhance the quality and clarity of the translated dialogue[176,177]. By integrating domain-specific knowledge from various sources, LLMs can generate more accurate and informative translations that cater to the needs of non-professional audiences.

Evaluating the performance of multilingual LLMs in medical language translation requires a multi-faceted approach. Some of the models use multiple choice question and answering test data with the calculation of accuracy score[174,175]. For generative benchmark, such as summarization[176,177], quantitative metrics such as BLEU[199], ROUGE[200], METEOR[201], and BERTScore[202] are commonly used to assess translation quality, but they should be supplemented with domain-specific evaluation criteria. For medical translations, accuracy of terminology, preservation of clinical meaning, and consistency across languages are crucial factors. Human evaluation by bilingual medical experts is essential to validate the nuanced understanding of medical concepts across languages. For patient-oriented translations, comprehension tests with lay individuals can assess the effectiveness of jargon simplification.

**Discussion** In both translation and simplification tasks, misinterpretation is a common occurrence that can have damaging consequences. In developing and deploying medical translation and simplification platforms, developers should prioritize professional datasets, such as textbooks and peer-reviewed journals for medical knowledge recall. This way, it will be less likely for misinformation from unreliable web sources to skew the output[213]. Another ethical consideration of using LLMs to perform medical translation is the potential for discriminatory verbiage to be inserted inadvertently into the output. Such verbiage is difficult to prevent due to the nature of the pipeline. This may cause miscommunications and even have legal consequences.[214].

### 4.6 Medical Education

LLMs can be incorporated into the medical education system in different ways, including facilitating study through explanations, aiding in language translation, answering questions, assisting with medical exam preparation, and providing Socratic-style tutoring[215,152]. Therefore, medical education could involve text generation, text simplification, semantic textual similarity, information retrieval, and etc. It has been suggested that medical education can be augmented by generating scenarios, problems, and corresponding answers by an LLM. Students will gain a richer educational experience through personalized study modules and case-based assessments, encountering a wider array of challenges and scenarios beyond those found in standard textbooks[214]. LLMs can also generate feedback on student responses to practical problems, allowing students to know their areas of weakness in real time. Inherently, these will better prepare these medical students for the real world since they would have been exposed to more scenarios[216].

Another use of LLMs in the medical field is educating the public. Medical dialogues are often complex and difficult to understand for the average patient. LLMs can tune the textual output of prompts to use varying degrees of medical terminology for different audiences. This will make medical information easy to understand for the average person while ensuring medical professionals have access to the most precise information[214].

**Guideline** Integrating LLMs into medical education can start with existing pre-trained models such as ChatGPT[217], and Med-Gemini[108]. Instead of developing models from scratch, it is often more effective to leverage the knowledge synthesis, question answering, and content generation capabilities of these powerful models. For instance, ChatGPT[178] can provide explanations and clarifications on complex medical concepts, facilitating self-study and reinforcing understanding. Med-Gemini[108], a multimodal model, can analyze medical images and generate detailed reports, aiding in the training of diagnostic skills. Institutions are exploring the integration of these language models into curricula, leveraging their strengths while ensuring proper oversight and ethical considerations. As this technology continues to advance, it holds promise for enhancing the efficiency and accessibility of medical education while complementing human expertise.

To evaluate the effectiveness of integrating LLMs into medical education, a combination of quantitative and qualitative methods should be employed. Current research focuses on the QA based evaluation[108]. Quantitative metrics can include student performance on assessments, such as exam scores and clinical skills evaluations, comparing outcomes before and after the introduction of LLM-based tools. Qualitative methods, such as surveys and focus groups, can gather feedback from students and educators on the perceived benefits, challenges, and areas for improvement in using LLMs for learning and teaching. Additionally, longitudinal studies can track the long-term impact of LLM integration on student learning outcomes, clinical competence, and career preparedness. By employing a comprehensive evaluation framework, institutions can iteratively refine their approach to leveraging LLMs in medical education, ensuring that these powerful tools are effectively harnessed to enhance learning while maintaining educational quality and ethical standards.

**Discussion** Potential downsides of using LLMs in medical education include the current lack of ethical training and biases in training datasets[24]. These biases, if not addressed, can propagate through the generated outputs, reinforcing stereotypes and potentially leading to discrimination in medical education. The lack of explicit ethical training during LLM development may also result in the generation of content that does not align with the ethical principles and guidelines of the medical profession, such as promoting unethical practices or violating patient privacy.

Furthermore, the risk of misinformation, particularly in the form of hallucinations, presents a challenge in utilizing LLMs for medical education. LLMs can generate plausible-sounding but factually incorrect information, which can mislead students and healthcare professionals if relied upon without proper verification. This can lead to the propagation of misconceptions, inappropriate treatment strategies, or misdiagnosis[218]. To mitigate these risks, it is essential to establish rigorous fact-checking and validation processes and emphasize the importance of critical thinking, evidence-based practice, and the verification of information from multiple reliable sources in medical education.

### 4.7 Mental Health Support

Mental health support involves both diagnosis and treatment. For example, depression is treated through a variety of psychotherapies, including cognitive behavior therapy, interpersonal psychotherapy, psychodynamic therapy, etc.[153]. Many of these techniques are primarily dominated by patient-doctor conversations, with lengthy treatment plans that are cost-prohibitive for many. The ability of LLMs to serve as conversation partners and companions may lower the barrier to entry for patients with financial or physical constraints[219], increasing the accessibility to mental health treatments[180]. There have been various research works and discussions on the effects of incorporating LLMs into the treatment plan[180,220,221].

The level of self-disclosure has a heavy impact on the effectiveness of mental health diagnosis and treatment. The degree of willingness to share has a direct impact on the diagnosis results and treatment plan. Studies have shown that patient willingness to discuss mental health-related topics with a robot is high[222,220]. Alongside the convenience and lower financial stakes, mental health support by LLMs has the potential to be more effective than human counterparts in many scenarios.

**Guideline** Development and deployment of LLMs targeted at mental health support can start with an existing LLM. Instead of pre-training or fine-tuning on general medical data, it is often better to use medical question and answer data as most of the LLM's work will be talking to the patient, which involves back-and-forth conversation in the format of question and answering[223]. PsyChat[179] is a client-centric LLM dialogue system that provides psychological support comprising five modules: client behavior recognition, counselor strategy selection, input packer, response generator, and response selection. Specifically, the response generator is fine-tuned with ChatGLM-6B with a vast dialogue dataset. Through both automatic and human evaluations, the system has demonstrated its effectiveness and practicality in real-life mental health support scenarios. ChatCounselor is designed to provide mental health support. It initializes from Vicuna and fine-tunes from an 8k size instruct-tuning dataset collected from real-world counseling dialogue examples[180]. Psy-LLM is an LLM aimed to be an assistive mental health tool to support the workflow of professional counselors, particularly to support those who might be suffering from depression or anxiety[223]. Another work presents a comprehensive evaluation of prompt engineering, few-shot, and fine-tuning techniques on multiple LLMs in the mental health domain[181]. The results reveal that fine-tuning on a variety of datasets can improve LLM's capability on multiple mental-health-specific tasks across different datasets simultaneously[181]. The work also releases their model Mental-Alpaca and Mental-FLAN-T5 as open-source LLMs targeted at multiple mental health prediction tasks[181]. Evaluating mental health-focused language models involves a multi-faceted approach that combines automated metrics and expert human assessment. Automated evaluations measure the relevance, coherence, and empathy of the generated responses using specialized metrics tailored to the mental health domain. Mental health professionals conduct human evaluations through simulated counseling sessions, assessing the clinical appropriateness and therapeutic potential of the models' responses. Recent research has introduced various evaluation frameworks that integrate tasks such as text generation (conversational response)[223], QA[180] and mental health prediction[181]. Liu et al.[180] prompt GPT-4 to compare ChatCounselor's responses with other models based on specific criteria and explanations. This multi-faceted approach provides researchers with a thorough understanding of the strengths and limitations of mental health-focused language models, enabling them to refine the models and develop more effective and reliable tools for mental health support.

**Discussion** Two of the most critical difficulties in employing LLMs for mental health support are the lack of emotional understanding and the risk of inappropriate or harmful responses[224]. LLMs, being language models, may struggle to fully grasp and respond to the complex emotional states and needs of individuals seeking mental health support. They may not be able to provide the same level of empathy and human connection that is crucial in therapeutic interactions.

Moreover, if not properly trained or controlled, LLMs may generate responses that are inappropriate, insensitive, or even harmful to individuals in vulnerable emotional states[225]. They may provide advice that is not grounded in evidence-based psychological practices or that goes against established mental health guidelines. Addressing these challenges requires rigorous training of LLMs in evidence-based practices, ethical considerations, and risk assessment protocols, as well as collaboration between mental health professionals and AI researchers.

### 4.8 Medical Inquiry and Response

The rapid advancement of LLMs also opens up new possibilities for improving healthcare delivery and patient care. LLMs, trained on vast amounts of medical knowledge, have the potential to understand and generate human-like text, making them suitable for tasks such as answering patient inquiries and assisting physicians in documentation[190,226]. As the demand for accessible and efficient healthcare services grows, researchers are exploring the use of medical LLMs to alleviate the burden on healthcare professionals and provide patients with reliable information and support. Therefore, medical inquiry and response could involve entity extraction, information retrieval, question answering.

**Guideline** Large language models can be effectively integrated into medical consultation systems to provide AI-powered assistance to healthcare professionals and enhance patient care. Instead of relying solely on rule-based algorithms or limited datasets, these systems leverage the vast knowledge and reasoning capabilities of LLMs to engage in diagnostic conversations and provide personalized recommendations. For example, Healthcare Copilot[190] combines dialogue, memory, and processing components to enable safe patient-LLM interactions, enhance conversations with historical data, and summarize consultations. Similarly, Google's Articulate Medical Intelligence Explore (AMIE)[189] employs a novel self-play-based simulated environment with automated feedback mechanisms, allowing the system to learn and adapt across diverse medical contexts. Another LLM-based diagnostic system[191] emulates the thought processes of experienced physicians and leverages reinforcement learning techniques to assist in disease screening, initial diagnoses, and the parsing of medical guidelines. These pioneering systems showcase the potential of medical LLMs in providing high-quality, AI-powered consultations and assisting physicians in their daily practice, while emphasizing the importance of rigorous testing, ethical oversight, and collaboration between medical experts and AI researchers to ensure their safe and responsible deployment. These systems showcase the potential of medical LLMs in providing high-quality, AI-powered medical consultations and assisting physicians in their daily practice.

Current evaluation of these systems often involves the calculation of metrics such as accuracy, precision, recall, and F1-score[189]. Additionally, some studies conduct multi-dimensional assessments of the models' performance, examining aspects such as inquiry capability, conversational fluency, response accuracy and safety using benchmarks and comparisons with human experts or well-established models like ChatGPT[190]. However, these metrics alone are not sufficient for a comprehensive real-world assessment. It is adviced that the evaluation of this should focus on the diagnostic accuracy, patient satisfaction, and adherence to medical guidelines[227].

**Discussion** However, there is still far from deploying them in the real-world healthcare system. Several challenges must be addressed before widespread deployment in real-world healthcare settings. One major concern is the potential for biased or inaccurate outputs, which could lead to improper medical advice or misdiagnosis[218]. Rigorous testing and validation across diverse patient populations and medical contexts are essential to ensure the reliability and generalizability of these systems. Additionally, the integration of medical LLMs into existing healthcare workflows and infrastructure may require substantial technical and organizational efforts. Privacy and security concerns surrounding patient data must also be carefully considered and addressed.

Furthermore, the development and deployment of medical LLMs raise important ethical and responsible AI considerations. Ensuring transparency, explainability, and accountability in the decision-making processes of these systems is crucial to maintaining trust and facilitating informed consent from patients[228,229]. The potential impact on the doctor-patient relationship and the role of human physicians in an AI-assisted healthcare setting must also be carefully examined. Ongoing collaboration between AI researchers, healthcare professionals, ethicists, and policymakers will be necessary to establish guidelines and best practices for the responsible development and deployment of medical LLMs in real-world healthcare settings.

## 5 Challenges

We address the challenges and discuss solutions to the adoption of LLMs in an array of medical applications.

### 5.1 Hallucination

Hallucination of LLMs refers to the phenomenon where the generated output contains inaccurate or nonfactual information. It can be categorized into intrinsic and extrinsic hallucinations[230,218]. Intrinsic hallucination generates outputs logically contradicting factual information, such as wrong calculations of mathematical formulas[218]. Extrinsic hallucination happens when the generated output cannot be verified, typical examples include LLMs 'faking' citations that do not exist or 'dodging' the question. When integrating LLMs into the medical domain, fluent but nonfactual LLM hallucinations can lead to the dissemination of incorrect medical information, causing misdiagnoses, inappropriate treatments, and harmful patient education. It is therefore vital to ensure the accuracy of LLM outputs in the medical domain.

**Potential Solutions** Current solutions to mitigate LLM hallucination can be categorized into training-time correction, generation-time correction, and retrieval-augmented correction. The first (i.e. training-time correction) adjusts model parameter weights, thus reducing the probability of generating hallucinated outputs. Its examples include factually consistent reinforcement learning[231] and contrastive learning[232]. The second (i.e. generation-time correction) adds a 'reasoning' process to the LLM inference to ensure reliability, using drawing multiple samples[233] or a confidence score to identify hallucination before the final generation. The third approach (i.e. retrieval-augmented correction) utilizes external resources to mitigate hallucination, for example, using factual documents as prompts[234] or chain-of-retrieval prompting technique[235].

### 5.2 Lack of Evaluation Benchmarks and Metrics

Current benchmarks and metrics often fail to evaluate LLM's overall capabilities, especially in the medical domain. For example, MedQA (USMLE)[14] and MedMCQA[148] offer extensive coverage on QA tasks but fail to evaluate important LLM-specific metrics, including trustworthiness, helpfulness, explainability, and faithfulness[206]. It is therefore imperative to develop domain and LLM-specific benchmarks and metrics.

**Potential Solutions** Singhal et al.[10] proposed HealthSearchQA consisting of commonly searched health queries, offering a more human-aligned benchmark for evaluating LLM's capabilities in the medical domain. Benchmarks such as TruthfulQA[236] and HaluEval[237] evaluate more LLM-specific metrics, such as truthfulness, but do not cover the medical domain. Future research is necessary to meet the need for more medical and LLM-specific benchmarks and metrics than what is currently available.

### 5.3 Domain Data Limitations

Current datasets in the medical domain (Table 2) remain relatively small compared to datasets for training general-purpose LLMs (Table 1). These limited small datasets only cover a small space[10] of the vase domain of medical knowledge. This results in LLMs exhibiting extraordinary performance on open benchmarks with extensive data coverage, yet falling short on real-life tasks such as differential diagnosis and personalized treatment planning[11].

Although the volume of medical and health data is large, most require extensive ethical, legal, and privacy procedures to be accessed. In addition, these data are often unlabeled, and solutions to leverage these data, such as human labeling and unsupervised learning[238], face challenges due to the lack of human expert resources and small margins of error.

**Potential Solutions** Current state-of-the-art approaches[11,15] typically fine-tune the LLMs on smaller open-sourced datasets to improve their domain-specific performance. Another solution is to generate high-quality synthetic datasets using LLMs to broaden the knowledge coverage; however, it has been discovered that training on generated datasets causes models to forget[239]. Future research is needed to validate the effectiveness of using synthetic data for LLMs in the medical field.

### 5.4 New Knowledge Adaptation

LLMs are trained on extensive data to learn knowledge. Once trained, it is expensive and inefficient to inject new knowledge into an LLM through re-training. However, it is sometimes necessary to update the knowledge of the LLM, for example, on a new adverse effect of a medication or a novel disease. Two problems occur during such knowledge updates. The first problem is how to make LLMs appropriately 'forget' the old knowledge, as it is almost impossible to remove all 'old knowledge' from the training data, and the discrepancy between new and old knowledge can cause unintended association and bias[240]. The second problem is the timeliness of the additional knowledge - how do we ensure the model is updated in real-time[241]? Both problems pose substantial barriers to using LLMs in medical fields, where accurate and timely updates of medical knowledge are crucial in real-world implementations.

**Potential Solutions** Current solutions to knowledge adaptation can be categorized into model editing and retrieval-augmented generation. Model editing[242] alters the knowledge of the model by modifying its parameters. However, this method does not generalize well, with their effectiveness varying across different model architectures. In contrast, retrieval-augmented generation provides external knowledge sources as prompts during model inference; for example, Lewis et al.[243] enabled model knowledge updates by updating the model's external knowledge memory.

### 5.5 Behavior Alignment

Behavior alignment refers to the process of ensuring that the LLM's behaviors align with the objectives of its task. Development efforts have been spent on aligning LLMs with general human behavior, but the behavior discrepancy between general humans and medical professionals remains challenging for adopting LLMs in the medical domain. For example, ChatGPT is well aligned with general human behavior, but their answers to medical consultations are not as concise and professional as those by human experts[45]. In addition, misalignment in the medical domain introduces unnecessary harm and ethical concerns[244] that lead to undesirable consequences.

**Potential Solutions** Current solutions include instruction fine-tuning, reinforcement learning from human feedback (RLHF)[45], and prompt tuning[132,129]. Instruction fine-tuning[124] refers to improving the performance of LLMs on specific tasks based on explicit instructions. For example, Ouyang et al.[45] used it to help LLMs generate less toxic and more suitable outputs. RLHF uses human feedback to evaluate and align the outputs of LLMs. It is effective in multiple tasks, including becoming helpful chatbots[245] and decision-making agents[246]. Prompt tuning can also align LLMs to the expected output format. For example, Liu et al.[247] uses a prompting strategy, chain of hindsight, to enable the model to detect and correct its errors, thus aligning the generated output with human expectations.

### 5.6 Ethical and Safety Concerns

Concerns have been raised regarding using LLMs (e.g., ChatGPT) in the medical domain[248], with a focus on ethics, accountability, and safety. For example, the scientific community has disapproved of using ChatGPT in writing biomedical research papers[228] due to ethical concerns. The accountability of using LLMs as assistants to practice medicine is challenging[123,249]. Li et al.[250] and Shen et al.[229] found that prompt injection can cause the LLM to leak personally identifiable information (PII), e.g., email addresses, from its training data, which is a substantial vulnerability when implementing LLM in the medical domain.

**Potential Solutions** With no immediate solutions available, we have nevertheless observed research efforts to understand the cause of these ethical and legal concerns. For example, Wei et al.[251] propose that PII leakage is attributed to the mismatched generalization between safety and capability objectives (i.e., the pre-training of LLMs utilizes a larger and more varied dataset compared to the dataset used for safety training, resulting in many of the model's capabilities are not covered by safety training).

### 5.7 Regulatory Challenges

The regulatory landscape of LLMs presents distinct challenges due to their large scale, broad applicability and varying reliability across applications. As LLMs progressively permeate the fields of medicine and healthcare, their versatility allows a single LLM family to facilitate a multitude of tasks across a broad spectrum of interest groups. This represents a substantial departure from the AI-based medical technologies of the past, which were typically tailored to meet specific medical needs and cater to particular interest groups[252,192]. In addition, the recent innovations of AI-enabled personalized approaches in areas such as oncology also present challenges to the traditional one-for-all auditing process[253]. This divergence and innovation necessitate regulators to develop adaptable, foresightful frameworks to ensure the safety, ethical standards, and privacy of the new family of LLMs-powered medical technologies.

**Potential Solutions** To address the complex regulatory challenges without hindering innovation, regulators should devise adaptive, flexible, and robust frameworks. Drawing on the insights from Mesko and Topol[252], creating a dedicated regulatory category and implementing patient design to enhance decision-making for LLMs used for medical purposes can better address their unique attributes and minimize harm. Furthermore, the insights outlined by Derraz et al.[253] emphasize the importance of implementing agile regulatory frameworks that can keep pace with the fast-paced advancements in personalized applications. Researchers both inside[252,253] and outside of healthcare[254,255] have proposed innovative strategies to regulate the use of LLMs involving (i) assessing LLMs-enabled applications in real-world settings, (ii) obligations of transparency of data and algorithms, (iii) adaptive risk assessment and mitigation processes, (iv) continuous testing and refinement of audited technologies. Such proactive regulatory adaptations are crucial to maintaining high standards of safety, ethics, and trustworthiness of medical technology.

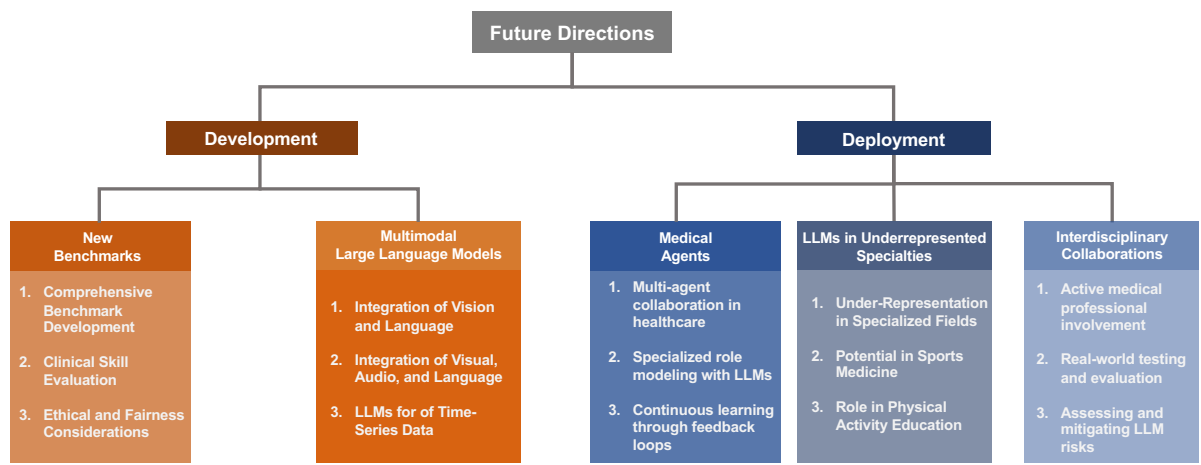

**Figure 6.** Future directions of LLMs in clinical medicine in terms of both development and deployment.

## 6 Future Directions

Although LLMs have already made an impact on people's lives through chatbots and search engines, their integration into medicine is still in the infant stage. As shown in Figure 6, numerous new avenues of medical LLMs await researchers and practitioners to explore how to better serve the general public and patients.

### 6.1 Introduction of New Benchmarks

Recent studies have underscored the shortcomings of existing benchmarks in evaluating LLMs for clinical applications [256,257]. Traditional benchmarks, which primarily gauge accuracy in medical question-answering, inadequately capture the full spectrum of clinical skills necessary for LLMs [10]. Criticisms have been leveled against the use of human-centric standardized medical exams for LLM evaluation, arguing that passing these tests does not necessarily reflect an LLM's proficiency in the nuanced expertise required in real-world clinical settings [10]. In response, there is an emerging consensus on the need for more comprehensive benchmarks. These should include capabilities like sourcing from authoritative medical references, adapting to the evolving landscape of medical knowledge, and clearly communicating uncertainties [19,10]. To further enhance the relevance of these benchmarks, new benchmarks should incorporate scenarios that test an LLM's ability through simulation of real-world applications and adjust to feedback from clinicians while maintaining robustness. Additionally, considering the sensitive nature of healthcare, these benchmarks should also assess factors such as fairness, ethics, and equity, which, though crucial, pose quantification challenges [10]. While efforts such as the AMIE study have advanced benchmarking by utilizing real physician evaluations and comprehensive criteria rooted in actual clinical skills and communication, as reflected in the Objective Structured Clinical Examination (OSCE), there remains a pressing need for benchmarks that are adaptive, scalable and robust for other diverse and personalized applications of LLMs. The aim is to create benchmarks that more effectively mirror diverse real-world clinical scenarios, thus providing a more accurate measure of LLMs' suitability for their applications in medicine. Future research may focus on (i) using synthetic data along with real-world data to create benchmarks that are both comprehensive and scalable, (ii) using clinical guidelines and criteria to reflect real-world values that are not normally included in traditional benchmarks, (iii) physician-in-the-loop benchmarks to evaluate the performance of LLMs leveraging their human counterparts or users.

### 6.2 Multimodal LLM Integrated with Time-Series, Visual, and Audio Data

Multimodal LLMs (MLLMs), or Large Multimodal Models (LMMs), are LLM-based models designed to perform multimodal (e.g., involving both visual and textual) tasks [258]. While LLMs primarily address NLP tasks, MLLMs support a broader range of tasks, such as comprehending the underlying meaning of a meme and generating website codes from images. This versatility suggests promising applications of MLLMs in medicine. Several MLLM-based frameworks integrating vision and language, e.g., MedPaLM M [259], LLaVA-Med [260], Visual Med-Alpaca [261], Med-Flamingo [262], and Qilin-Med-VL [263], have been proposed to adopt the medical image-text pairs for fine-tuning, thus enabling the medical LLMs to efficiently understand the input medical (e.g., radiology) images. A recent study [264] proposes to integrate vision, audio, and language inputs for automated diagnosis in dentistry. However, there exist only very few medical LLMs that can process time series data, such as electrocardiograms (ECGs)[265] and sphygmomanometers (PPGs)[266], despite such data being important for medical diagnosis and monitoring. Although early in their proposed research stages, these studies suggest that MLLMs trained at scale have the potential to effectively generalize across various domains and modalities outside of NLP tasks. However, the training of MLLMs at scale is still costly and ineffective, resulting in the size of MLLMs being much smaller than LLMs. Moving forward, future research may focus on (i) more effective processing, representation, and learning of multi-modal data and knowledge, (ii) cost-effective training of MLLMs, especially modalities that are more resource-demanding such as videos and images, (iii) collecting or accessing safely, currently unavailable, multi-modal data in medicine and healthcare.

### 6.3 Medical Agents

LLM-based agents[267,268] utilize LLMs as controllers to leverage their reasoning capabilities. By integrating LLMs with external tools and multimodal perceptions, these agents can interact with environments, learn from feedback, and acquire new skills, enabling them to solve complex tasks (e.g., software design, molecular dynamics simulation) through human-like behaviors, such as role-playing and communication[269,270].

However, integrating these agents effectively within the medical domain remains a challenge. The medical field involves numerous roles[270] and decision-making processes, especially in disease diagnosis that often requires a series of investigations involving CT scans, ultrasounds, electrocardiograms, and blood tests. The idea of utilizing LLMs to model each of these roles, thereby creating collaborative medical agents, presents a promising direction. These agents could mimic the roles of radiologists, cardiologists, pathologists, etc., each specializing in interpreting specific types of medical data. For example, a radiologist agent could analyze CT scans, while a pathologist agent could focus on blood test results. The collaboration among these specialized agents could lead to a more holistic and accurate diagnosis. By leveraging the comprehensive knowledge base and contextual understanding capabilities of LLMs, these agents not only interpret individual medical reports but also integrate these interpretations to form a cohesive medical opinion. To enhance the integration of LLMs-based agents, future research may explore (i) a seamless data pipeline that collects data from various devices and transforms them into data format compatible with LLMs (ii) effective communication and collaboration between agents, especially in areas such as ensuring truthfulness during communication, dispute resolution between agents, and role-based data security measures, (iii) real-time decision-making such as making timely decisions using data collected from remote monitoring devices, (iv) adaptive learning such as preparing for a new pandemic or learning from unseen medical conditions.

### 6.4 LLMs in Underrepresented Specialties

Current LLM research in medicine has largely focused on general medicine, likely due to the greater availability of data in this area[11,249]. This has resulted in the under-representation of LLM applications in specialized fields like 'rehabilitation therapy' or 'sports medicine'. The latter, in particular, holds potential, given the global health challenges posed by physical inactivity. The World Health Organization identifies physical inactivity as a major risk factor for non-communicable diseases (NCDs), impacting over a quarter of the global adult population[271]. Despite initiatives to incorporate physical activity (PA) into healthcare systems, implementation remains challenging, particularly in developing countries with limited PA education among healthcare providers[271]. LLMs could play a pivotal role in these settings by disseminating accurate PA knowledge and aiding in the creation of personalized PA programs[272]. Such applications could enhance PA levels, improving global health outcomes, especially in resource-constrained environments. To spark innovation in these underrepresented specialties, future research can focus on areas such as (i) effective data collection in underrepresented specialties, (ii) applications of LLMs in assisting with tasks of underrepresented specialties, (iii) using LLMs to help progress the research of these underrepresented specialties.

### 6.5 Interdisciplinary Collaborations

Just as interdisciplinary collaborations are crucial in safety-critical areas like nuclear energy production, collaborations between the medical and technology communities for developing medical LLMs are essential to ensure AI safety and efficacy in medicine. The medical community has primarily adopted LLMs provided by technology companies without rigorously questioning their data training, ethical protocols, or privacy protection. Medical professionals are therefore encouraged to actively participate in creating and deploying medical LLMs by providing relevant training data, defining the desired benefits of LLMs, and conducting tests in real-world scenarios to evaluate these benefits[19,21,22]. Such assessments would help to determine the legal and medical risks associated with LLM use in medicine and inform strategies to mitigate LLM hallucination[273]. Additionally, training 'bilingual' professionals—those versed in both medicine and LLM technology—is increasingly vital due to the rapid integration of LLMs in healthcare. Future research may explore (i) interdisciplinary frameworks, such as frameworks to facilitate the sharing of localized data from rural clinics, (ii) 'bilingual education programs' that offer training from both worlds - AI and medicine, (iii) effective in-house development methods to help hospitals and physicians 'guard' patient data from corporations while still being able to embrace innovation.

## Acknowledgements

This work was supported in part by the Pandemic Sciences Institute at the University of Oxford; the National Institute for Health Research (NIHR) Oxford Biomedical Research Centre (BRC); an NIHR Research Professorship; a Royal Academy of Engineering Research Chair; the Well-come Trust funded VITAL project; the UK Research and Innovation (UKRI); the Engineering and Physical Sciences Research Council (EPSRC); and the InnoHK Hong Kong Centre for Cerebro-cardiovascular Engineering (COCHE).

## Author Contributions

FL, ZL, JL, and DC supervised the project. FL conceived and designed the study. HZ, FL, BG, XZ, JH, and WJ conducted the literature review, performed data analysis, and drafted the manuscript. All authors contributed to the interpretation and final manuscript preparation. All authors read and approved the final manuscript.

## Competing Interests

The authors declare no competing interests.